\documentclass{article}

\usepackage{microtype}
\usepackage{graphicx}
\usepackage{subcaption}
\usepackage{booktabs} 

\usepackage{hyperref}

\usepackage[preprint]{icml2026}

\usepackage{amsmath}
\usepackage{amssymb}
\usepackage{mathtools}
\usepackage{amsthm}

\usepackage{algorithm}
\usepackage{algorithmic}
\usepackage{svg}

\usepackage[capitalize,noabbrev]{cleveref}

\theoremstyle{plain}

\theoremstyle{definition}

\theoremstyle{remark}

\usepackage[textsize=tiny]{todonotes}

\icmltitlerunning{Kill it with FIRE: On Leveraging Latent Space Directions for Runtime Backdoor Mitigation in Deep Neural Networks}

\begin{document}

\twocolumn[
  \icmltitle{Kill it with FIRE: On Leveraging Latent Space Directions for Runtime Backdoor Mitigation in Deep Neural Networks}



  \icmlsetsymbol{equal}{*}

  \begin{icmlauthorlist}
    \icmlauthor{Enrico Ahlers}{hub}
    \icmlauthor{Daniel Passon}{hub}
    \icmlauthor{Yannic Noller}{rub}
    \icmlauthor{Lars Grunske}{hub}
  \end{icmlauthorlist}

  \icmlaffiliation{hub}{Humboldt University of Berlin, Germany}
  \icmlaffiliation{rub}{Ruhr University Bochum, Germany}

  \icmlcorrespondingauthor{Enrico Ahlers}{enrico.philip.ahlers@hu-berlin.de}

  \icmlkeywords{Machine Learning, ICML}

  \vskip 0.3in
]



\printAffiliationsAndNotice{}  

\begin{abstract}

Machine learning models are increasingly present in our everyday lives; as a result, they become targets of adversarial attackers seeking to manipulate the systems we interact with.
A well-known vulnerability is a \textit{backdoor} introduced into a neural network by poisoned training data or a malicious training process. Backdoors can be used to induce unwanted behavior by including a certain trigger in the input.
Existing mitigations filter training data, modify the model, or perform expensive input modifications on samples. If a vulnerable model has already been deployed, however, those strategies are either ineffective or inefficient.
To address this gap, we propose our inference-time backdoor mitigation approach called \textit{FIRE} (\emph{Feature-space Inference-time REpair}). We hypothesize that a trigger induces structured and repeatable changes in the model's internal representation. We view the trigger as \textit{directions} in the latent spaces between layers that can be applied in reverse to correct the inference mechanism. Therefore, we turn the backdoored model against itself by manipulating its latent representations and moving a poisoned sample's features along the backdoor directions to neutralize the trigger.
Our evaluation shows that \textit{FIRE} has low computational overhead and outperforms current runtime mitigations on image benchmarks across various attacks, datasets, and network architectures.
\end{abstract}

\section{Introduction}

Deep neural networks are widely used across real-world disciplines such as computer vision \cite{krizhevskyImageNetClassificationDeep2012, simonyanDeepConvolutionalNetworks2014}, autonomous driving \cite{tampuuSurveyEndtoEndDriving2022}, healthcare \cite{litjensSurveyDeepLearning2017} and finance \cite{mienyeDeepLearningFinance2024}. Their deployment in safety- and security-critical applications, however, has raised significant concerns regarding their robustness and trustworthiness, particularly in adversarial settings.

One prominent threat to the integrity of deep neural networks is the presence of backdoors. Backdoor attacks deliberately manipulate the training data or training procedure by embedding hidden triggers that cause the model to exhibit attacker-chosen behavior at inference time \cite{guBadNetsEvaluatingBackdooring2019,  zengNarcissusPracticalCleanLabel2023,mengaraBackdoorAttacksDeep2024, liShortcutsEverywhereNowhere2026,wuDefensesAdversarialMachine2026}. While the model performs normally on benign inputs, the presence of a specific trigger can reliably induce misclassification into a target label. Due to their stealthy nature and high attack success rates, backdoors pose a serious risk in scenarios where training is outsourced, or pre-trained models are obtained from untrusted sources.

As a result, substantial research effort has been devoted to the detection and mitigation of backdoors in neural networks. Existing defenses can be categorized into pre-training, training, post-training, deployment, and inference-time approaches \cite{wuDefensesAdversarialMachine2026}. Pre-training and training-time defenses require access to the data or training process and are therefore inapplicable once a model has been deployed, while post-training and deployment-time methods often involve modifying model weights or parameters, or rejecting suspicious inputs.

In this work, we focus on inference-time defenses, which aim to mitigate backdoor behavior during runtime without retraining, finetuning or modifying the weights of the model. Existing inference-time approaches often either act as firewalls that detect and reject suspected inputs or attempt to remove triggers through input modification. A key limitation of firewall-style defenses is that simply discarding flagged samples is not acceptable in many safety-critical settings where a decision is required. For example, if a backdoored traffic sign classifier encounters a traffic sign with an adversarial sticker, the system cannot ignore the sign; it must recover the correct prediction to ensure safe behavior. Existing input modification methods can sometimes enable such correction, but they often incur substantial computational cost and are slow at inference time \cite{shiBlackboxBackdoorDefense2023, yangSampDetoxBlackboxBackdoor2024}. These limitations motivate lightweight inference-time mitigation methods that correct poisoned predictions without modifying the model parameters.

In our approach, we take a different perspective by analyzing the internal latent representations of backdoored models. We show that examining multiple latent spaces provides valuable insight into how backdoors manifest within the network. In particular, we study the directions induced by the trigger in latent space and discuss approaches for identifying these directions. Our analysis reveals that poisoned samples align along consistent directions in certain latent spaces of the network.
The particular latent spaces in which this behavior emerges depend on the attack method, model architecture, and dataset.

Building on this observation, we propose a novel runtime backdoor mitigation method that leverages the latent directions associated with the trigger. Instead of modifying the input image or changing the model parameters, our approach operates directly on the latent representation of the input during inference. By moving the representation along the identified backdoor directions, the poisoned sample is effectively purified, restoring the correct prediction. This procedure does not require expensive optimization or access to large clean datasets and can be applied at runtime without altering the network. Owing to its low computational overhead, the mitigation can be updated online during inference, allowing the estimation of the backdoor directions to improve progressively as new samples are processed.

Extensive experiments across multiple datasets and network architectures demonstrate that the proposed method consistently outperforms existing runtime mitigation techniques. These results suggest that latent-space analysis offers a powerful and practical avenue for understanding and defending against backdoor attacks, enabling efficient and non-intrusive mitigation in deployed deep learning systems.

\section{Preliminaries and Related Work}

\subsection{Backdoor Attack}
Backdoor attacks constitute a class of training-time attacks in which an adversary embeds hidden triggers into a model by poisoning the training data or manipulating the training procedure \cite{guBadNetsIdentifyingVulnerabilities2019,bagdasaryanHowBackdoorFederated2020, xiGraphBackdoor2021}. When the trigger is present at inference time, the model reliably produces an attacker-chosen target prediction while maintaining high accuracy on clean inputs \cite{nguyenWaNetImperceptibleWarpingbased2020}. Backdoor attacks can be highly effective and generalize across modalities (e.g., vision, speech, text, and reinforcement learning) while evading standard validation, including via clean-label poisoning, adaptive/input-aware triggers, and transfer- or pretraining-based injection \cite{turnerLabelConsistentBackdoorAttacks2019,  nguyenInputAwareDynamicBackdoor2020, kuritaWeightPoisoningAttacks2020,zhaiBackdoorAttackSpeaker2021, xuInstructionsBackdoorsBackdoor2024}.
\subsection{Backdoor Mitigation}
In response, a wide range of defense mechanisms has been proposed across different stages of the machine learning pipeline \cite{jinSurveyBackdoorAttacks2025,wuDefensesAdversarialMachine2026}.
Among post-training defenses, existing methods typically require either 
(i) training additional components (e.g., auxiliary modules or purification generators) \cite{doanFebruusInputPurification2020,zhuNeuralPolarizerLightweight2023,alkaderhammoudDontFREAKOut2023,liBackdoorMitigationCorrecting2023,huangOrionOnlineBackdoor2023,xuReliableEfficientBackdoor2023,luoSifterNetGeneralizedModelAgnostic2025,zhuClassConditionalNeuralPolarizer2025, chenREFINEInversionFreeBackdoor2025}, (ii) solving expensive nontrivial optimization problems at deployment (e.g., UNIT \cite{chengUNITBackdoorMitigation2024} and MMAC \cite{wangImprovedActivationClipping2024}), or (iii) modifying the deployed model via parameter updates \cite{liNeuralAttentionDistillation2020,usmanNNrepairConstraintbasedRepair2021}, pruning \cite{liuFinePruningDefendingBackdooring2018}, or unlearning \cite{wangNeuralCleanseIdentifying2019,wangPatchBasedBackdoorsDetection2022}. However, defenders often lack access to the required datasets or computational resources, motivating inference-time techniques.

Inference-time defenses are particularly relevant for deployed systems and include detection-based approaches, such as SentiNet \cite{chouSentiNetDetectingLocalized2020} and STRIP \cite{gaoSTRIPDefenceTrojan2019}, which aim to identify poisoned inputs and then discard suspicious samples. However, in settings like autonomous driving, abstention is infeasible, and the prediction must instead be corrected.

For inference-time correction, input-space purification methods have shown promising results.
A lightweight baseline is ShrinkPad \cite{liBackdoorAttackPhysical2021}, though diffusion-based purification can achieve higher accuracy. ZIP \cite{shiBlackboxBackdoorDefense2023} applies a pretrained diffusion model entirely at test time, but is comparatively expensive at inference. In contrast, SampDetox \cite{yangSampDetoxBlackboxBackdoor2024} requires a domain-specific diffusion model, and BDFirewall \cite{liBDFirewallEffectiveExpeditiously2025} additionally trains a U-Net on top of such a model, violating our \emph{no-training} constraint. Other defenses further require mask localization and/or iterative inpainting optimization \cite{usmanRuleBasedRuntimeMitigation2022,weiLightweightBackdoorDefense2023,maySalientConditionalDiffusion2023,sunMaskRestoreBlind2025}, which increases runtime cost and can depend on spatial trigger structure, which limits they applicability; BDMAE \cite{sunMaskRestoreBlind2025} is designed for localized triggers; Shi et al. show it breaks down on image-wide triggers, which our setting requires handling \yrcite{shiBlackboxBackdoorDefense2023}.

\subsection{Latent-Space Representations}
\label{sec:latent_analysis}

We use \emph{feature space} and \emph{latent space} interchangeably to refer to intermediate representations in a deep neural network. A substantial body of work has shown that the latent spaces of deep neural networks encode rich semantic structure \cite{bankAutoencoders2023,huComplexityMattersRethinking2023, jebreelDefendingBackdoorAttacks2024}. In generative models, especially GANs, linear and non-linear directions in latent space have been shown to correspond to high-level semantic attributes, enabling controlled manipulation of generated images \cite{pariharEverythingThereLatent2022}. Similar observations have been made for discriminative models, where internal representations capture class-relevant and concept-level information. Recent work demonstrates that attributes, styles, and semantic variations can often be isolated along specific latent directions, sometimes consistently across samples \cite{pariharEverythingThereLatent2022}. These findings suggest that latent spaces provide a meaningful and structured representation of input data. Importantly, prior analysis shows the backdoor signal is not guaranteed to be most separable in the final (commonly used) latent layer; the strongest benign–poisoned feature difference can occur at an earlier critical layer \cite{jebreelDefendingBackdoorAttacks2024}. This work builds on the insight that backdoor behavior is reflected in internal representations and leverages latent directions as a practical mechanism for inference-time backdoor mitigation.

To the best of our knowledge, \textit{FIRE} is the first method to use latent-space representations for inference-time sample correction, leveraging them to update its correction as new samples arrive.

\subsection{Network and Data Notation}
Let a feed-forward neural network $f$ be a composition of $L$ layers
\begin{equation}
f \;=\; f_{L-1} \circ f_{L-2} \circ \cdots \circ f_0,
\end{equation}
where each layer is a mapping between Euclidean spaces
\begin{equation}
f_\ell:\mathbb{R}^{d_{\ell-1}} \to \mathbb{R}^{d_\ell}, \qquad \ell \in \{0,\dots,L-1\},
\end{equation}
with input dimension $d_{-1}$ and output dimension $d_{L-1}$ (e.g.\ logits).
For any input $x\in\mathbb{R}^{d_{-1}}$, define the depth-$\ell$ latent map $h_\ell:\mathbb{R}^{d_{-1}}\to\mathbb{R}^{d_\ell}$ and the corresponding tail network $g_\ell:\mathbb{R}^{d_\ell}\to\mathbb{R}^{d_{L-1}}$ by
\begin{equation}
h_\ell \;=\; f_\ell \circ f_{\ell-1} \circ \cdots \circ f_0,\qquad h_0\;\coloneqq\;f_0,
\end{equation}
and
\begin{equation}
g_\ell \;\coloneqq\; f_{L-1} \circ f_{L-2} \circ \cdots \circ f_{\ell+1}, \qquad g_{L-1}\;\coloneqq\;\mathrm{id} 
\end{equation}
so that $f(x)=g_\ell\!\big(h_\ell(x)\big)$ for all $\ell$.

\paragraph{Clean and Poisoned Samples}
Let $\{x_i^{\mathrm{clean}}\}_{i=1}^N$ be $N$ clean samples in $\mathbb{R}^{d_{-1}}$.
A \emph{trigger injection operator} is a (possibly input-dependent) function $\mathcal{T}:\mathbb{R}^{d_{-1}}\to\mathbb{R}^{d_{-1}}$ that adds a trigger to an input. The poisoned counterpart is
\begin{equation}
x_i^{\mathrm{pois}} \;\coloneqq\; \mathcal{T}\!\big(x_i^{\mathrm{clean}}\big).
\label{eq:add_trigger_input}
\end{equation}

\paragraph{Latent Displacements}

For each paired sample
\(
(x_i^{\mathrm{clean}}, x_i^{\mathrm{pois}})
\)
we denote their representations after layer $f_\ell$ by
\begin{equation}
x_{i,\ell}^{\mathrm{clean}} \;\coloneqq\; h_\ell\!\big(x_i^{\mathrm{clean}}\big)\in\mathbb{R}^{d_\ell},
\end{equation}
\begin{equation}
x_{i,\ell}^{\mathrm{pois}} \;\coloneqq\; h_\ell\!\big(x_i^{\mathrm{pois}}\big)\in\mathbb{R}^{d_\ell}.
\end{equation}
We then define the \emph{sample-wise trigger displacement} in latent space after layer $f_{\ell}$ as
\begin{equation}
b_{i,\ell} \;\coloneqq\; x_{i,\ell}^{\mathrm{pois}} - x_{i,\ell}^{\mathrm{clean}} \in \mathbb{R}^{d_\ell}.
\label{eq:samplewise_backdoor_displacement}
\end{equation}

Importantly, $b_{i,\ell}$ is allowed to depend on the sample index $i$ and (implicitly) on the trigger mechanism $\mathcal{T}$; i.e.\ displacements can vary across inputs and triggers.

\section{Methodology}

\subsection{Intuition and Observations}
Semantic manipulations can be captured as approximately linear directions in latent representations, as shown by FLAME~\cite{pariharEverythingThereLatent2022}. Building on this perspective, we next estimate a \emph{general} trigger that can be interpreted as an (approximately) linear direction in one or multiple latent spaces of the network.

In Eq.~\ref{eq:samplewise_backdoor_displacement}, we defined the \emph{sample-wise} trigger displacement $b_{i,\ell}$.
Empirically, we show that backdoored models often encode the individual displacements $b_{i,\ell}$ in a highly consistent manner in at least one latent space, i.e., these vectors exhibit strong alignment (or low-variance structure) across many samples. Based on this perspective, we introduce \textit{FIRE} (\emph{Feature-space Inference-time REpair}), an inference-time mitigation approach that corrects poisoned predictions by intervening directly in intermediate representations.

This observation is compatible with the hierarchical nature of learned representations: earlier layers tend to capture low-level and coarse visual primitives, while deeper layers represent increasingly complex and task-specific features~\cite{zeilerVisualizingUnderstandingConvolutional2014}. Consequently, the latent space(s) in which a general trigger direction emerges depends on the trigger's complexity: simple patterns may manifest as a consistent shift in earlier representations, whereas more structured triggers may only align in deeper latent spaces.

Once such a general direction has been identified, the trigger effect can be modeled as an additive component in latent space. This enables a simple repair operation: adding the direction applies the trigger in latent space, while subtracting it removes (or reduces) the trigger contribution before forwarding the corrected representation through the remaining network.

To summarize, \textit{FIRE} is based on the hypothesis that backdoors induce a structured, repeatable displacement in latent space that becomes sufficiently coherent at some depth to be approximated by a linear direction.
\subsection{Estimating the General Trigger Direction}
\label{sec:estimate_trigger_and_layer}
We propose two strategies to obtain latent directions, depending on the defender's knowledge; additional strategies are provided in Appendix~\ref{app:strategies}.

\paragraph{Strategy 1: Paired Clean/Poisoned Samples}
If paired samples are available, i.e., we have access to both a clean and a corresponding poisoned version of the same underlying input, we can directly form pairs $(x_i^{\mathrm{clean}}, x_i^{\mathrm{pois}})$. If paired samples are not available, but the trigger injection operator $\mathcal{T}$ is known (e.g., via trigger reconstruction), we can instead generate pairs by applying $\mathcal{T}$ to clean inputs, as in Eq.~\ref{eq:add_trigger_input}.
For a fixed candidate layer $f_\ell$, we compute the sample-wise displacement in latent space,
\begin{equation}
b_{i,\ell}
\;=\;
h_{\ell}\!\big(x_i^{\mathrm{pois}}\big) - h_{\ell}\!\big(x_i^{\mathrm{clean}}\big) 
\in \mathbb{R}^{d_\ell},
\end{equation}
which corresponds to Eq.~\ref{eq:samplewise_backdoor_displacement}. Repeating this for $N$ pairs yields a general direction by averaging:
\begin{equation}
\widehat{b}_{\ell}
\;\coloneqq\;
\frac{1}{N}\sum_{i=1}^N b_{i,\ell}
\;=\;
\frac{1}{N}\sum_{i=1}^N
\Big(
h_{\ell}\!\big(x_i^{\mathrm{pois}}\big) - h_{\ell}\!\big(x_i^{\mathrm{clean}}\big)
\Big).
\label{eq:backdoor_direction_mean}
\end{equation}
This estimator directly captures the dominant shared component of the per-sample displacements at layer $f_\ell$.

\paragraph{Strategy 2: Unpaired Clean and Poisoned Sets (Unknown Trigger) with Image Augmentations}
In many settings, the trigger in input space is unknown, which makes it infeasible to construct paired samples. Instead, we assume access to an unpaired set of clean samples $\{x_i^{\mathrm{clean}}\}_{i=1}^{N_c}$ and a set of poisoned samples $\{x_j^{\mathrm{pois}}\}_{j=1}^{N_p}$ (not necessarily corresponding to the same underlying images). For a candidate layer $f_\ell$, we compute empirical centroids in latent space,
\begin{align}
\widehat{\mu}^{\mathrm{clean}}_{\ell}
&\;\coloneqq\;
\frac{1}{N_c}\sum_{i=1}^{N_c} h_{\ell}\!\big(x_i^{\mathrm{clean}}\big),
\label{eq:clean_centroid_offline}
\\
\widehat{\mu}^{\mathrm{pois}}_{\ell}
&\;\coloneqq\;
\frac{1}{N_p}\sum_{j=1}^{N_p} h_{\ell}\!\big(x_j^{\mathrm{pois}}\big),
\end{align}
and define a centroid-difference of the trigger direction as
\begin{equation}
\widehat{b}_{\ell}^{\mathrm{diff}}
\;\coloneqq\;
\widehat{\mu}^{\mathrm{pois}}_{\ell} - \widehat{\mu}^{\mathrm{clean}}_{\ell}.
\label{eq:centroid_trigger_direction}
\end{equation}
If, after layer $f_\ell$, poisoned representations exhibit an approximately consistent shift relative to clean ones, then $\widehat{b}_{\ell}^{\mathrm{diff}}$ averages out sample-specific variation across examples and leaves the common shift direction induced by the trigger.
 
\noindent\textbf{Augmentation-based estimate}
Gao et al.\ report that backdoor-trigger features tend to be more fragile under image corruptions than genuine semantic features~\yrcite{gaoEnergybasedBackdoorDefense2024}. We exploit this by constructing pairs consisting of a poisoned input and an augmented copy of the same poisoned input. Let $\mathcal{A}:\mathbb{R}^{d_{-1}}\to\mathbb{R}^{d_{-1}}$ denote an augmentation/corruption operator (e.g., additive noise, blur, JPEG compression, random resizing/cropping), and define
\begin{equation}
x_i^{\mathrm{pois,aug}} \;\coloneqq\; \mathcal{A}\!\big(x_i^{\mathrm{pois}}\big).
\end{equation}
For a candidate layer $f_\ell$, we compute the sample-wise augmentation-induced displacement in latent space,
\begin{equation}
b^{\mathrm{aug}}_{i,\ell}
\;\coloneqq\;
h_{\ell}\!\big(x_i^{\mathrm{pois}}\big) - h_{\ell}\!\big(x_i^{\mathrm{pois,aug}}\big)
\in \mathbb{R}^{d_\ell}.
\label{eq:aug_samplewise_displacement}
\end{equation}
Under the assumption that semantic features are comparatively stable under $\mathcal{A}$ while trigger-related features degrade disproportionately, $b^{\mathrm{aug}}_{i,\ell}$ is dominated by the trigger change and thus serves as a proxy for the backdoor direction. Aggregating over $N_p$ poisoned samples yields
\begin{equation}
\widehat{b}_{\ell}^{\mathrm{aug}}
\;\coloneqq\;
\frac{1}{N_p}\sum_{i=1}^{N_p} b^{\mathrm{aug}}_{i,\ell}.
\label{eq:aug_trigger_direction_mean}
\end{equation}

\noindent\textbf{Combined direction}
Finally, we combine the centroid-difference and augmentation-based estimates using trade-off parameter $\lambda$ to obtain a single trigger direction at layer $f_\ell$:
\begin{equation}
\widehat{b}_{\ell}
\;\coloneqq\;
\lambda\,\widehat{b}_{\ell}^{\mathrm{diff}} + (1-\lambda)\,\widehat{b}_{\ell}^{\mathrm{aug}},
\qquad \lambda \in [0,1].
\label{eq:combined_trigger_direction}
\end{equation}

\subsection{Direction-Based Latent Repair}
\label{sec:remove-trigger}
Given an estimated general trigger direction $\widehat{b}_\ell$ at layer $f_\ell$, we repair an input ${x}$ by modifying its latent representation $x_\ell \coloneqq h_\ell({x})$ such that the trigger contribution is reduced. Conceptually, we move $x_\ell$ in the opposite direction of $\widehat{b}_\ell$, i.e., we ``subtract'' the trigger component before forwarding the corrected representation through the tail network $g_\ell$.

The repair operator subtracts the direction estimate,
\begin{equation}
\tilde{x}_\ell
\;=\;
\mathrm{Rep}_\ell(x_\ell, \widehat{b}_\ell, \alpha_\ell)
\;\coloneqq\;
x_\ell - \alpha_\ell \widehat{b}_\ell,
\label{eq:rep_subtract}
\end{equation}
where $\alpha_\ell\in\mathbb{R}$ controls repair strength (typically $\alpha_\ell=1$).

The repaired prediction is obtained by forwarding $\tilde{x}_\ell$ through the rest of the network, i.e., $y_{\mathrm{rep}} = \arg\max g_\ell(\tilde{x}_\ell)$.

\begin{figure*}[t]
  \centering

  \begin{subfigure}[t]{0.305\textwidth}
    \centering
    \includegraphics[width=\linewidth]{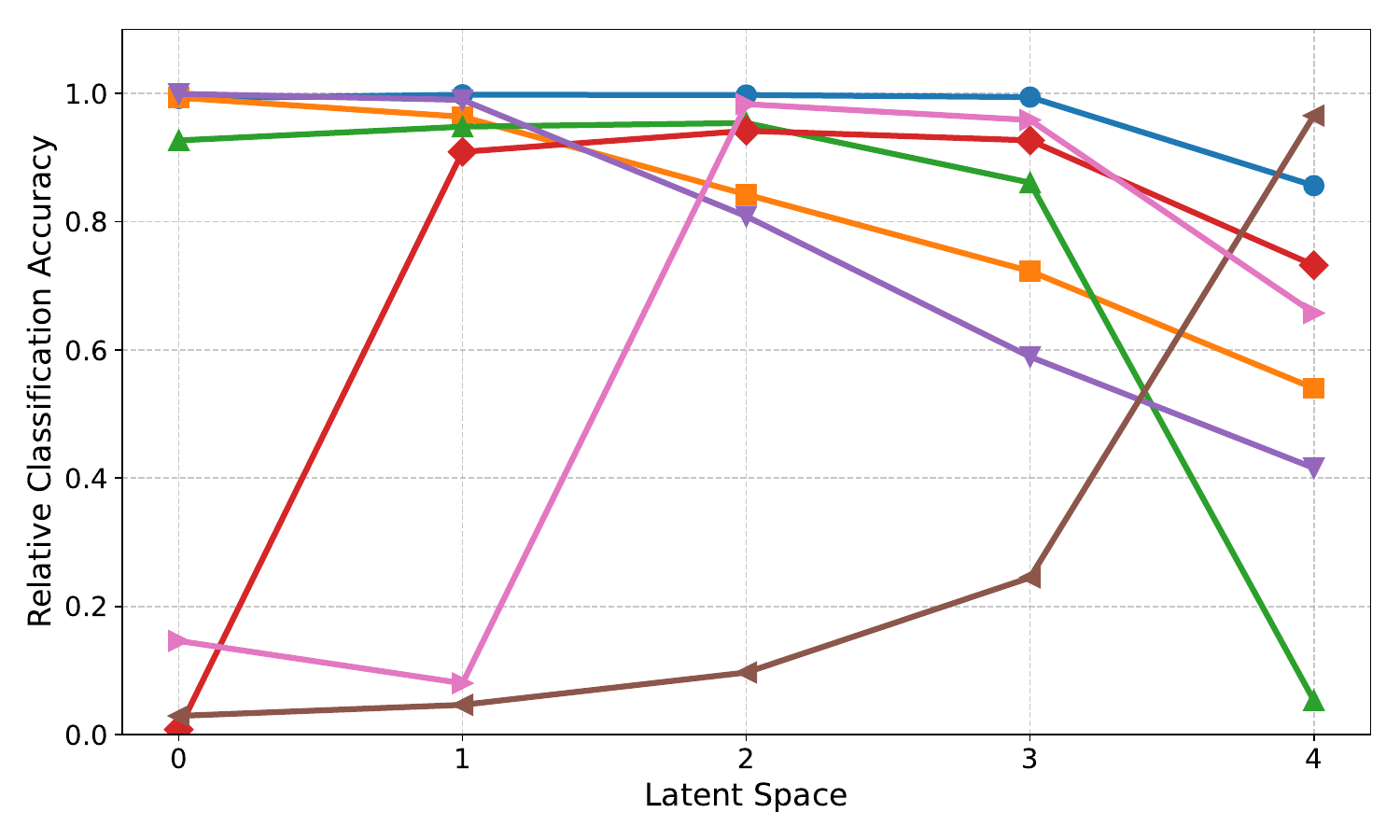}
    \caption{CIFAR-10 with PreActResNet18}
    \label{fig:fire_latent_cifar}
  \end{subfigure}\hfill
  \begin{subfigure}[t]{0.305\textwidth}
    \centering
    \includegraphics[width=\linewidth]{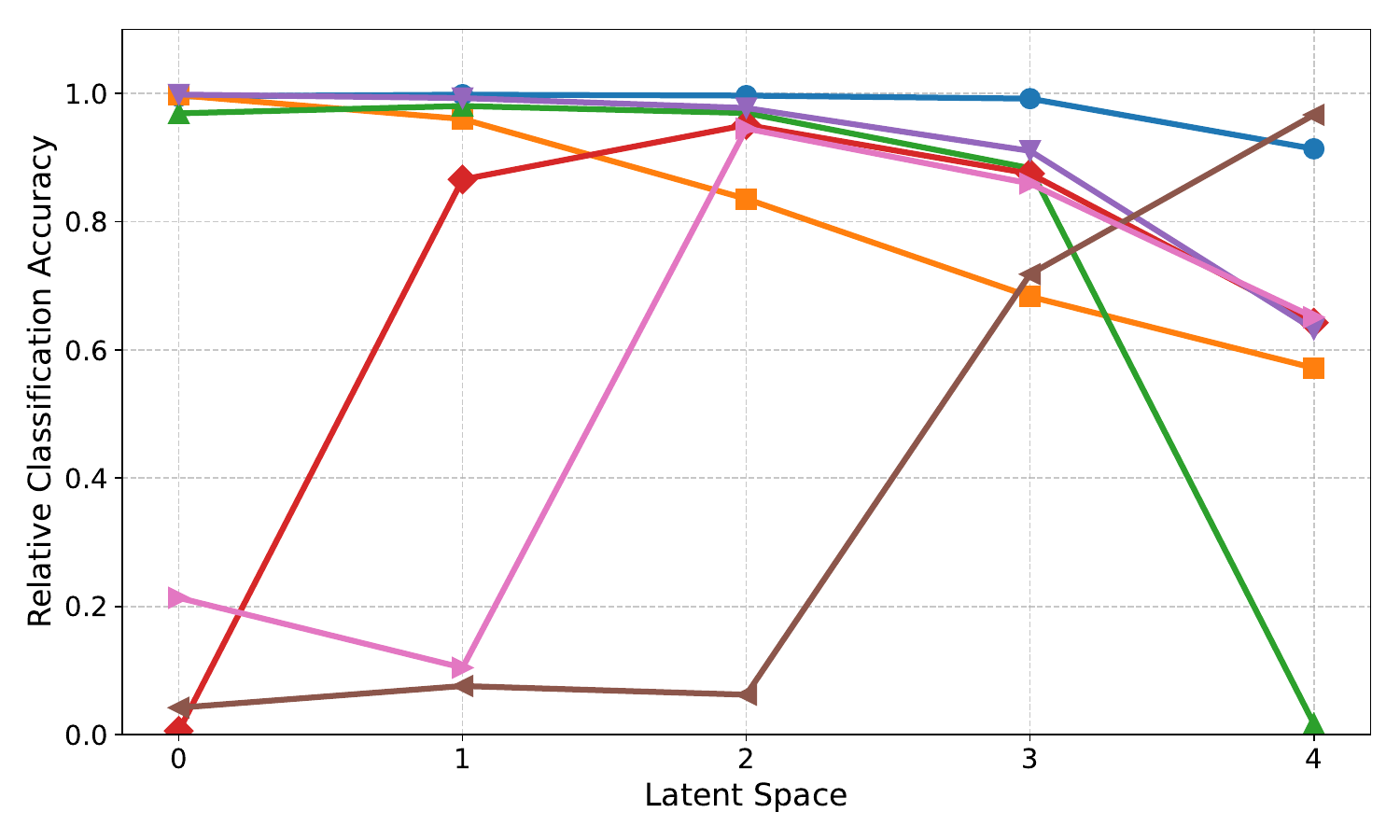}
    \caption{CIFAR-10 with PreActResNet34}
    \label{fig:fire_latent_res34}
  \end{subfigure}
  \begin{subfigure}[t]{0.38\textwidth}
    \centering
    \includegraphics[width=\linewidth]{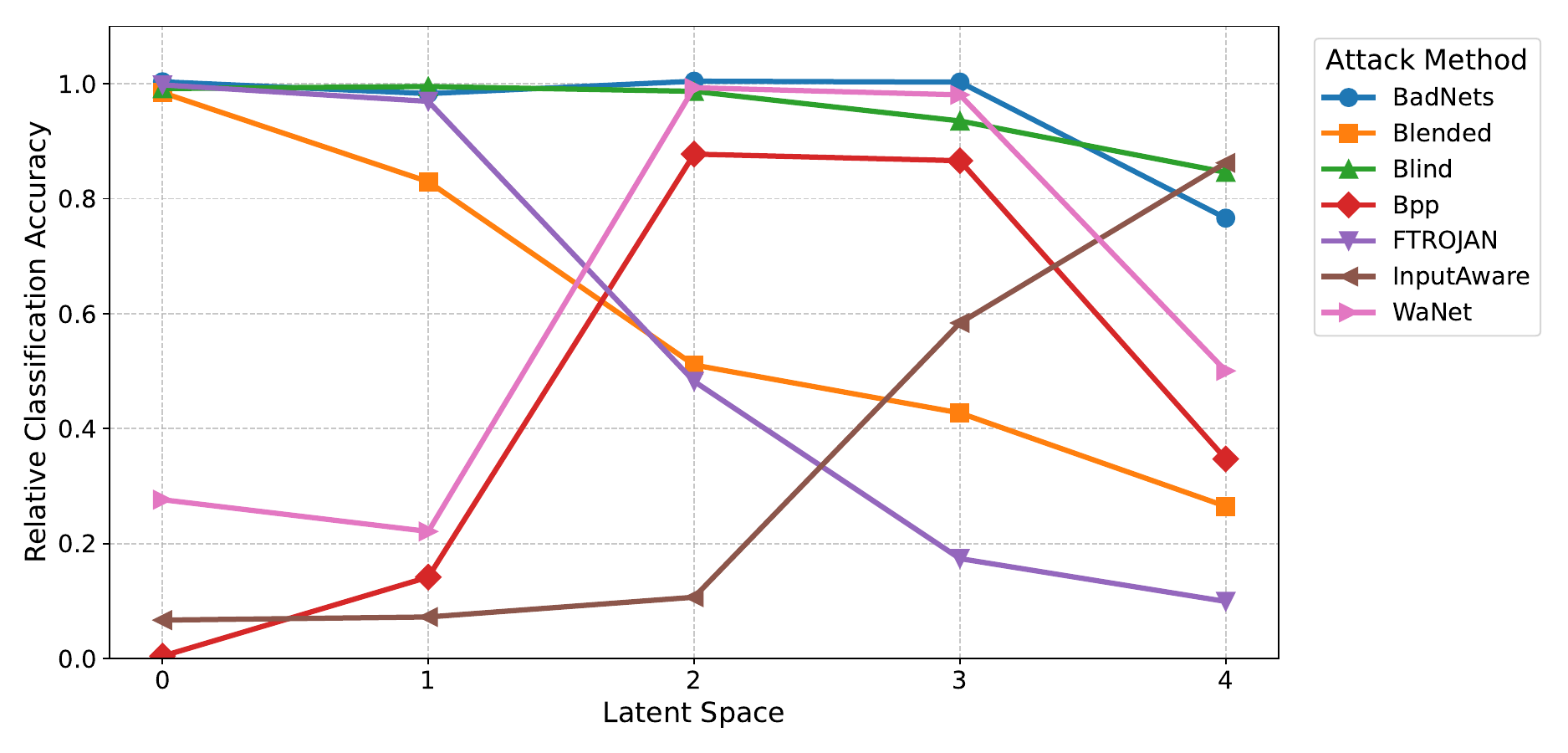}
    \caption{GTSRB with PreActResNet18}
    \label{fig:fire_latent_gtsrb}
  \end{subfigure}\hfill

  \caption{Mitigation performance when intervening \emph{manually} at each latent space. The x-axis shows the selected latent space, while the y-axis shows the \emph{relative} classification accuracy (\%), i.e. the Poisoned Accuracy (PA) divided by the Clean Accuracy (CA) of the model.}
  \label{fig:fire_latent_all}
\end{figure*}

\section{Empirical Study}
\label{sec:proof-of-concept}
We now empirically test the hypothesis from Section~3: in a backdoored network, the trigger induces a consistent, approximately linear displacement in latent space at certain depths. If such a direction can be identified, subtracting it at inference time should reduce the trigger effect and recover the clean prediction. A key focus is \emph{layer dependence}: we analyze at which layers the displacement is sufficiently coherent for latent-space repair to be effective.

To isolate the effect of latent space selection from estimation noise, we construct $N=100$ paired samples $(x_i^{\mathrm{clean}},x_i^{\mathrm{pois}})$ by inserting known triggers in input space and estimate a direction $\widehat{b}_\ell$ for each latent space using Strategy~1 (Eq.~\ref{eq:backdoor_direction_mean}).

\paragraph{Configurations}
\label{par:configs}
We evaluate on CIFAR-10 \cite{krizhevskyLearningMultipleLayers2009} and GTSRB \cite{stallkampGermanTrafficSign2011} using PreActResNet classifiers \cite{heIdentityMappingsDeep2016} trained under a poisoning ratio of $10\%$ across a range of backdoor attacks: BadNets \cite{guBadNetsIdentifyingVulnerabilities2019}, Blended \cite{chenTargetedBackdoorAttacks2017}, Blind \cite{bagdasaryanBlindBackdoorsDeep2021}, Bpp \cite{wangBppAttackStealthyEfficient2022}, FTROJAN \cite{wangInvisibleBlackBoxBackdoor2022}, InputAware \cite{nguyenInputAwareDynamicBackdoor2020}, and WaNet \cite{nguyenWaNetImperceptibleWarpingbased2020}. 5 latent spaces are used; one after the first convolutional layer, and one after each block.

We benchmark mitigation by intervening at each candidate latent space after layer $f_\ell$. Concretely, for each $\ell$ we compute the repaired prediction
\begin{equation}
y^{(\ell)}_{\mathrm{rep}}(x)
\;=\;
\arg\max\; g_\ell\!\Big(\mathrm{Rep}_\ell\big(h_\ell(x)\big)\Big),
\end{equation}
and report the Poisoned Accuracy (PA, \%), defined as the accuracy on poisoned inputs with respect to the ground-truth labels, as a function of $\ell$.

Fig.~\ref{fig:fire_latent_all} summarizes the resulting \emph{relative} layer-wise performance distribution across attacks. The key takeaway is that repair is highly depth-dependent: certain attacks become reliably correctable only after the trigger effect emerges as a coherent displacement in deeper representations, while others admit effective repair in earlier latent spaces. Intuitively, different attacks introduce different trigger features (e.g., localized patches or full-image warps), and these features become separable and stable at different representational stages. For example, the WaNet attack can be best mitigated at latent space 2, while a mitigation at latent space 0 and 1 yields poor results. Notably, these preferred intervention depths are highly consistent across different datasets and model architectures, suggesting that the optimal depth is driven primarily by the backdoor attack mechanism, rather than dataset- or architecture-specific effects.

Tab.~\ref{tab:aggregated_results} reports the correction performance of poisoned images to their true labels at different latent spaces. Across all three dataset--architecture configurations, we observe consistently strong recovery of the true label in at least one latent space. In many cases, the PA after correction is close to the Clean Accuracy (CA). The lowest performance occurs for the InputAware attack on a PreActResNet18 on GTSRB, where PA reaches 84.1\% compared to a CA of 97.51\%, i.e., a gap of 13.41\%. For over half of the configurations, we identify a direction that recovers the true label within 1\% of the CA.

\begin{table*}[]
\caption{Correction performance on 21 configurations, reporting Clean Accuracy (CA, \%) and Poisoned Accuracy (PA, \%) without defense, and Poisoned Accuracy (PA, \%) with \textit{FIRE} applied in different latent spaces. ``Diff to CA'' (\%) indicates the difference between CA and PA (CA -- PA) in the best-performing latent space.}
\label{tab:aggregated_results}
\vskip 0.1in
\begin{center}
\begin{small}
\begin{sc}
\begin{tabular}{llcc|ccccc|c}
\toprule
 & & & & \multicolumn{5}{c|}{PA in latent space $\uparrow$} & \\
Configuration & Attack & CA $\uparrow$ & PA $\uparrow$  & 0 & 1 & 2 & 3 & 4 & Diff to CA $\downarrow$ \\
\midrule
 & BadNets & 91.84 & 5.27 & 91.12 & \textbf{91.66} & 91.63 & 91.33 & 78.64 & 0.18 \\
 & Blended & 93.57 & 0.18 & \textbf{92.97} & 90.18 & 78.83 & 67.64 & 50.55 & 0.60 \\
 & Blind & 80.17 & 0.40 & 74.30 & 76.02 & \textbf{76.48} & 69.02 & 4.27 & 3.69 \\
\smash{\raisebox{-0.5\normalbaselineskip}{\shortstack{PreActResNet18 \\ CIFAR-10}}} & Bpp & 90.99 & 0.39 & 0.72 & 82.71 & \textbf{85.70} & 84.34 & 66.62 & 5.29 \\
 & FTROJAN & 93.56 & 0.00 & \textbf{93.52} & 92.63 & 75.64 & 55.13 & 38.91 & 0.04 \\
 & InputAware & 91.28 & 2.54 & 2.66 & 4.24 & 8.84 & 22.36 & \textbf{88.13} & 3.15 \\
 & WaNet & 91.19 & 3.90 & 13.33 & 7.30 & \textbf{89.69} & 87.42 & 59.96 & 1.50 \\
\midrule
 & BadNets & 96.05 & 3.73 & 96.39 & 94.41 & \textbf{96.49} & 96.34 & 73.60 & -0.44 \\
 & Blended & 97.05 & 0.27 & \textbf{95.56} & 80.52 & 49.57 & 41.47 & 25.70 & 1.49 \\
 & Blind & 57.28 & 41.23 & 56.81 & \textbf{57.02} & 56.54 & 53.58 & 48.47 & 0.27 \\
\smash{\raisebox{-0.5\normalbaselineskip}{\shortstack{PreActResNet18 \\ GTSRB}}} & Bpp & 96.79 & 0.24 & 0.40 & 13.71 & \textbf{84.96} & 83.84 & 33.62 & 11.84 \\
 & FTROJAN & 98.11 & 0.00 & \textbf{97.92} & 95.12 & 47.32 & 17.06 & 9.74 & 0.18 \\
 & InputAware & 97.51 & 6.17 & 6.52 & 7.05 & 10.40 & 56.93 & \textbf{84.10} & 13.41 \\
 & WaNet & 96.80 & 7.06 & 26.75 & 21.40 & \textbf{96.14} & 94.94 & 48.45 & 0.66 \\
\midrule
 & BadNets & 92.47 & 4.31 & 92.06 & \textbf{92.30} & 92.17 & 91.73 & 84.49 & 0.17 \\
 & Blended & 94.04 & 0.21 & \textbf{93.79} & 90.30 & 78.58 & 64.25 & 53.78 & 0.25 \\
 & Blind & 89.36 & 0.03 & 86.60 & \textbf{87.62} & 86.64 & 78.96 & 1.52 & 1.74 \\
\smash{\raisebox{-0.5\normalbaselineskip}{\shortstack{PreActResNet34 \\ CIFAR-10}}} & Bpp & 92.21 & 0.28 & 0.53 & 79.84 & \textbf{87.69} & 80.69 & 59.25 & 4.52 \\
 & FTROJAN & 93.60 & 0.06 & \textbf{93.45} & 92.93 & 91.48 & 85.22 & 59.18 & 0.15 \\
 & InputAware & 91.57 & 3.71 & 3.84 & 6.94 & 5.70 & 65.76 & \textbf{88.52} & 3.05 \\
 & WaNet & 90.15 & 9.74 & 19.24 & 9.40 & \textbf{85.22} & 77.49 & 58.69 & 4.93 \\
\bottomrule
\end{tabular}
\end{sc}
\end{small}
\end{center}
\vskip -0.1in
\end{table*}

\paragraph{Comparison to Prior Work}
Unlike \citet{jebreelDefendingBackdoorAttacks2024}, who emphasize that backdoor effects become most distinctive in the \emph{second half} of the network, we find that for \emph{purification} the optimal intervention depth is often much earlier: attacks such as Blended and BadNets are best mitigated in early latent spaces, while others (e.g., InputAware) still prefer later layers. This supports the view that repair works best at the earliest depth where the trigger is already encoded as a stable shift, before salient image content is discarded by later representations.

\section{Runtime Backdoor Mitigation}
\label{sec:runtime_mitigation}

\subsection{Threat Model and Deployment Setting}
\label{sec:threat_model}

We consider a defense scenario in which a potentially backdoored model is already trained and deployed, and the defender must mitigate malicious behavior \emph{during inference} under strict real-time constraints.

\paragraph{Attacker Capabilities}
The attacker has full control over the training pipeline. Concretely, the attacker can poison the training dataset and influence the training procedure itself, with the goal of implanting a backdoor into the resulting model. As in standard backdoor settings, the backdoored model behaves normally on benign inputs but, when a trigger is present, predicts an attacker-chosen target label with high probability.

\paragraph{Defender Capabilities}
The defender is assumed to have access to:
\begin{itemize}
    \item A \emph{small} set of clean reference samples (without triggers), denoted by $\mathcal{D}_{\mathrm{clean}} = \{x_i\}_{i=1}^{N_c}$.
    \item Inference access to the deployed model $f = f_{L-1} \circ \cdots \circ f_0$, including the ability to compute predictions $f(x)$ and to extract latent representations $h_\ell(x)$ for selected $\ell$ during a forward pass.
\end{itemize}

However, the defender \emph{cannot} assume access to the original training data, cannot modify model weights, and cannot afford computationally intensive procedures such as retraining, fine-tuning, or optimization-based purification at test time. The defense must therefore be a lightweight procedure implementable as an extension of the forward pass.

\paragraph{Runtime Input Stream}
\label{sec:input_stream}
At deployment, we consider a stream of flagged inputs $\{x_t\}_{t=1}^{\infty}$ that are treated as poisoned. Such a flagged stream can arise by filtering a larger, mostly clean deployment stream through an upstream backdoor-detection module (e.g., SCALE-UP~\cite{guoSCALEUPEfficientBlackbox2022} or related methods~\cite{wuDefensesAdversarialMachine2026})\footnote{We study the impact of imperfect detectors in Appendix~\ref{app:imperfect_detection}.
}. We assume the system cannot discard flagged inputs; in safety-critical settings, abstention may be infeasible and a prediction must be produced for every query. We therefore focus on \emph{post-detection mitigation}: efficiently recovering correct predictions on flagged inputs, analogous to the streaming-style runtime repair setting discussed in ZIP~\cite{shiBlackboxBackdoorDefense2023}.

\paragraph{Practical Motivation}
A canonical example is autonomous driving. Consider a backdoored traffic sign classifier deployed on a vehicle. If an attacker places a trigger (e.g., a small sticker) on a traffic sign, the classifier may output an incorrect label. The vehicle cannot ignore the sign or stop to run expensive defenses; it must instead correct the prediction \emph{immediately} to ensure safe behavior. This motivates a defense that operates directly on internal representations during a forward pass, without updating model weights.

\subsection{Defense Procedure}
\label{sec:runtime_defense_procedure}
\begin{figure*}[]
    \centering
    \includegraphics[width=1\linewidth]{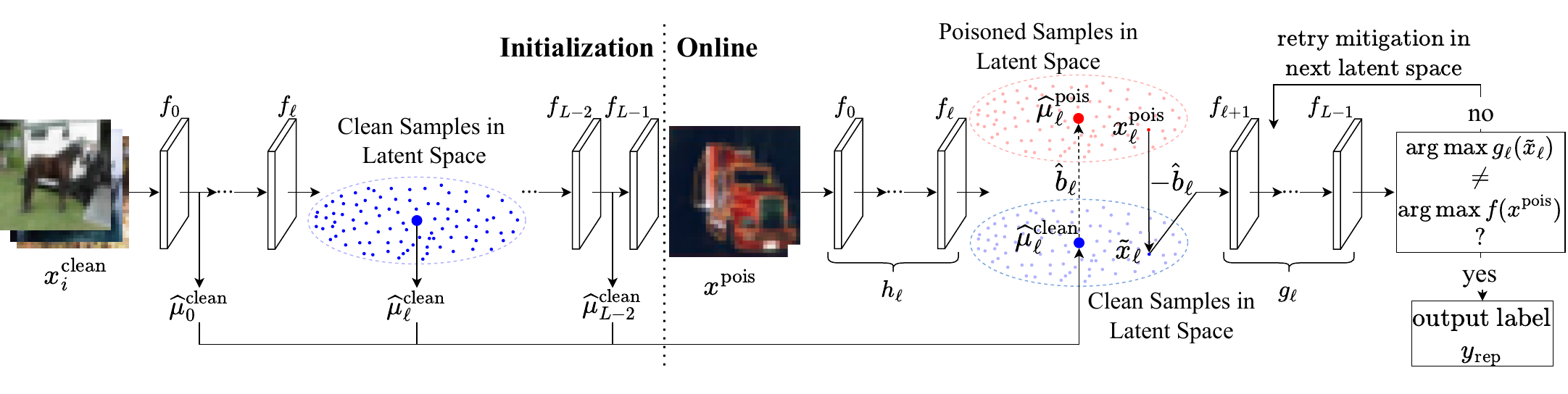}
    \caption{Overview of the defense procedure: The left side shows the short initialization phase, where a small set of clean samples is used to obtain the centroids $\widehat{\mu}^{\mathrm{clean}}_{\ell}$ in candidate latent spaces. The right side shows the online phase, where it is used to repair the latent representation of poisoned samples.}
    \label{fig:process_visualization}
\end{figure*}
We now instantiate the methodology from Section~\ref{sec:estimate_trigger_and_layer} and \ref{sec:remove-trigger} for the real-time threat model in Section~\ref{sec:threat_model}.
Alg.~\ref{alg:runtime_mitigation} summarizes the full streaming defense: it maintains online estimates of a trigger direction in multiple candidate latent spaces and applies repair at the earliest latent space that changes the model’s prediction on the current input.

\paragraph{Initialization}
As illustrated in Fig. \ref{fig:process_visualization}, we first compute a clean reference statistic for each candidate latent space after $f_\ell, \ell\in\mathcal{L}$, namely the clean centroid $\widehat{\mu}^{\mathrm{clean}}_\ell$ (Eq.~\ref{eq:clean_centroid_offline}). This provides a layer-wise baseline that online estimates are compared to.

\paragraph{Online Runtime Loop}
At runtime, the defense processes a stream of poisoned inputs. For each input, it first records the model’s unmitigated prediction and then evaluates candidate latent spaces in ascending order. At each latent space, the defense (i) updates its current estimate of the trigger direction using the arriving sample (Eq.~\ref{eq:centroid_trigger_direction}), (ii) repairs the corresponding latent representation (Eq.~\ref{eq:rep_subtract}), and (iii) checks whether this repaired representation alters the prediction. The defense returns the output label of the earliest mitigation attempt that alters the prediction; if no shift induces a change, it leaves the prediction unchanged. Because the trigger-direction estimates are updated continuously as the stream progresses, the defense improves over time and becomes progressively more effective as evidence accumulates in online estimates.

\section{Experiments}
\label{sec:experiments}
\subsection{Experimental Settings}
\label{sec:exp_settings}

Our implementation builds on BackdoorBench \cite{wuBackdoorBenchComprehensiveBenchmark2025}.
We evaluate on the same configurations as in Section~\ref{par:configs}; additional details are provided in Appendix~\ref{app:training_details}.
For our defense, we assume access to a small clean reference set and use $N_c=100$ clean samples by default to compute clean statistics and to support direction estimation\footnote{See Appendix~\ref{app:clean_samples} for results with different $N_c$.}. Strategy~2, with color jitter and Gaussian blurring as augmentation techniques, is used to obtain the latent directions. Furthermore, we choose the mixing weight $\lambda$ as 0.5 and the repair intensity parameter $\alpha$ as 1.

Lightweight inference-time methods that \emph{correct} predictions (rather than rejecting inputs) are relatively rare. We therefore compare against the state-of-the-art approach that supports runtime prediction correction: ZIP \cite{shiBlackboxBackdoorDefense2023} leverages diffusion-based purification and can correct predictions at test time, but incurs substantial computational cost.

\begin{algorithm}[t]
\caption{Runtime latent-space backdoor mitigation}
\label{alg:runtime_mitigation}
\begin{algorithmic}
  \STATE \textbf{Input:} model $f$, candidate layer indices $\mathcal{L}$, clean set $\mathcal{D}_{\mathrm{clean}}$, augmentation $\mathcal{A}$, mixing weight $\lambda\in[0,1]$
  \STATE Compute $\widehat{\mu}^{\mathrm{clean}}_\ell$ from $\mathcal{D}_{\mathrm{clean}}$ for all $\ell\in\mathcal{L}$ (Eq.~\ref{eq:clean_centroid_offline})
  \STATE Initialize $\widehat{\mu}^{\mathrm{pois}}_{\ell}\leftarrow 0$, $\widehat{\mu}^{\mathrm{pois,aug}}_{\ell}\leftarrow 0$ for all $\ell\in\mathcal{L}$
  \STATE Initialize counters $n_\ell \leftarrow 0$ and shifts $\widehat{b}_{\ell}\leftarrow 0$ for all $\ell\in\mathcal{L}$

  \FOR{each incoming poisoned sample $x_t^{\mathrm{pois}}$}
    \STATE $y\leftarrow \arg\max f(x_t^{\mathrm{pois}})$
    \STATE $x_t^{\mathrm{pois,aug}} \leftarrow \mathcal{A}\big(x_t^{\mathrm{pois}}\big)$
    \FOR{each $\ell\in\mathcal{L}$ in ascending order}
      \STATE $x_{t,\ell}^{\mathrm{pois}} \leftarrow h_\ell(x_t^{\mathrm{pois}})$,\quad $x_{t,\ell}^{\mathrm{pois,aug}} \leftarrow h_\ell(x_t^{\mathrm{pois,aug}})$,\quad $n_\ell \leftarrow n_\ell + 1$
      \STATE $\widehat{\mu}^{\mathrm{pois}}_\ell \leftarrow \widehat{\mu}^{\mathrm{pois}}_\ell + \frac{1}{n_\ell}\!\left(x_{t,\ell}^{\mathrm{pois}} - \widehat{\mu}^{\mathrm{pois}}_\ell\right)$
      \STATE $\widehat{\mu}^{\mathrm{pois,aug}}_\ell \leftarrow \widehat{\mu}^{\mathrm{pois,aug}}_\ell + \frac{1}{n_\ell}\!\left(x_{t,\ell}^{\mathrm{pois,aug}} - \widehat{\mu}^{\mathrm{pois,aug}}_\ell\right)$ 
      \STATE $\widehat{b}_\ell \leftarrow \lambda(\widehat{\mu}^{\mathrm{pois}}_\ell - \widehat{\mu}^{\mathrm{clean}}_\ell)+(1-\lambda)(\widehat{\mu}^{\mathrm{pois}}-\widehat{\mu}^{\mathrm{pois,aug}}_\ell)$
      \STATE $\tilde{x}_{t,\ell} \leftarrow \mathrm{Rep}_\ell\!\left(x_{t,\ell}^{\mathrm{pois}},\widehat{b}_\ell,1\right)$
      \STATE $y_{\mathrm{rep}} \leftarrow \arg\max g_\ell(\tilde{x}_{t,\ell})$
      \IF{$y_{\mathrm{rep}} \neq y$}
        \STATE \textbf{Output} $y_{\mathrm{rep}}$ and \textbf{break}
      \ENDIF
    \ENDFOR
    \STATE If no layer changed the prediction, \textbf{output} $y$
  \ENDFOR
\end{algorithmic}
\end{algorithm}

\begin{table*}[t]
\caption{Benchmark results for the streaming scenario: Clean Accuracy (CA, \%) and Poisoned Accuracy (PA, \%, higher is better) without a defense, Poisoned Accuracy (PA, \%) for the ZIP and \textit{FIRE} approaches, and the Time in milliseconds (ms, lower is better) are reported.}
\label{tab:runtime_results}
\begin{center}
\begin{small}
\begin{sc}
\begin{tabular}{llcc|ccc|cll}
\toprule
Configuration & Attack & CA $\uparrow$ & PA $\uparrow$ & ZIP PA $\uparrow$ & \multicolumn{2}{c}{FIRE PA $\uparrow$} & ZIP Time $\downarrow$ & \multicolumn{2}{c}{FIRE Time $\downarrow$} \\
\cmidrule(lr){6-7} \cmidrule(lr){8-8} \cmidrule(lr){9-10}
 & & & & & Pos 1 & Pos 10 & Online & Init. & Online \\
\midrule
 & BadNets & 91.86 & 5.27 & 27.19 & 22.56 & \textbf{86.22} & 1408.0 & 223.7 & \textbf{14.0} \\
 & Blended & 93.57 & 0.18 & 45.00 & 25.44 & \textbf{64.00} & 1403.3 & 222.0 & \textbf{12.1} \\
 & Blind & 80.20 & 0.40 & 0.55 & 3.78 & \textbf{69.78} & 1418.0 & 228.5 & \textbf{15.2} \\
\smash{\raisebox{-0.5\normalbaselineskip}{\shortstack{PreActResNet18 \\ CIFAR-10}}} & Bpp & 91.07 & 0.39 & 63.24 & 29.44 & \textbf{78.22} & 1404.4 & 213.6 & \textbf{11.8} \\
 & FTROJAN & 93.56 & 0.00 & 61.02 & 29.44 & \textbf{67.67} & 1398.6 & 220.4 & \textbf{12.0} \\
 & InputAware & 91.29 & 2.54 & 40.23 & 24.56 & \textbf{72.11} & 1424.4 & 214.4 & \textbf{12.1} \\
 & WaNet & 91.24 & 3.90 & 33.63 & 17.56 & \textbf{71.44} & 1452.3 & 219.2 & \textbf{16.8} \\
\midrule
 & BadNets & 96.05 & 3.73 & 6.25 & 76.53 & \textbf{86.95} & 1460.6 & 225.8 & \textbf{11.7} \\
 & Blended & 97.06 & 0.27 & 31.25 & 34.61 & \textbf{61.89} & 1472.8 & 231.8 & \textbf{12.8} \\
 & Blind & 57.33 & 41.23 & 27.73 & 21.88 & \textbf{49.01} & 1504.3 & 225.5 & \textbf{11.5} \\
\smash{\raisebox{-0.5\normalbaselineskip}{\shortstack{PreActResNet18 \\ GTSRB}}} & Bpp & 96.80 & 0.24 & 69.92 & \textbf{85.28} & 82.66 & 1485.9 & 228.2 & \textbf{11.7} \\
 & FTROJAN & 98.12 & 0.00 & 67.58 & \textbf{79.16} & 76.37 & 1494.6 & 241.0 & \textbf{12.0} \\
 & InputAware & 97.53 & 6.17 & 50.39 & \textbf{71.20} & 61.89 & 1531.6 & 232.6 & \textbf{12.2} \\
 & WaNet & 96.79 & 7.06 & 31.25 & 20.29 & \textbf{42.32} & 1474.9 & 226.2 & \textbf{14.2} \\
\midrule
 & BadNets & 92.46 & 4.31 & 24.80 & 26.78 & \textbf{88.67} & 1449.7 & 208.5 & \textbf{22.1} \\
 & Blended & 94.02 & 0.21 & 45.66 & 26.89 & \textbf{61.78} & 1454.3 & 205.5 & \textbf{20.4} \\
 & Blind & 89.43 & 0.03 & 0.74 & 19.78 & \textbf{78.44} & 1398.4 & 212.1 & \textbf{24.2} \\
\smash{\raisebox{-0.5\normalbaselineskip}{\shortstack{PreActResNet34 \\ CIFAR-10}}} & Bpp & 92.23 & 0.28 & 63.40 & 25.44 & \textbf{71.11} & 1407.8 & 209.4 & \textbf{21.8} \\
 & FTROJAN & 93.63 & 0.06 & 62.34 & 29.11 & \textbf{82.89} & 1433.4 & 210.8 & \textbf{21.1} \\
 & InputAware & 91.61 & 3.71 & 52.19 & 21.11 & \textbf{71.33} & 1409.6 & 221.1 & \textbf{23.2} \\
 & WaNet & 90.17 & 9.74 & 12.46 & 13.33 & \textbf{50.78} & 1409.5 & 209.5 & \textbf{26.2} \\
\bottomrule
\end{tabular}
\end{sc}
\end{small}
\end{center}
\end{table*}

\subsection{Runtime Mitigation Performance}
\begin{figure}[t]
  \centering
  \includegraphics[width=\columnwidth]{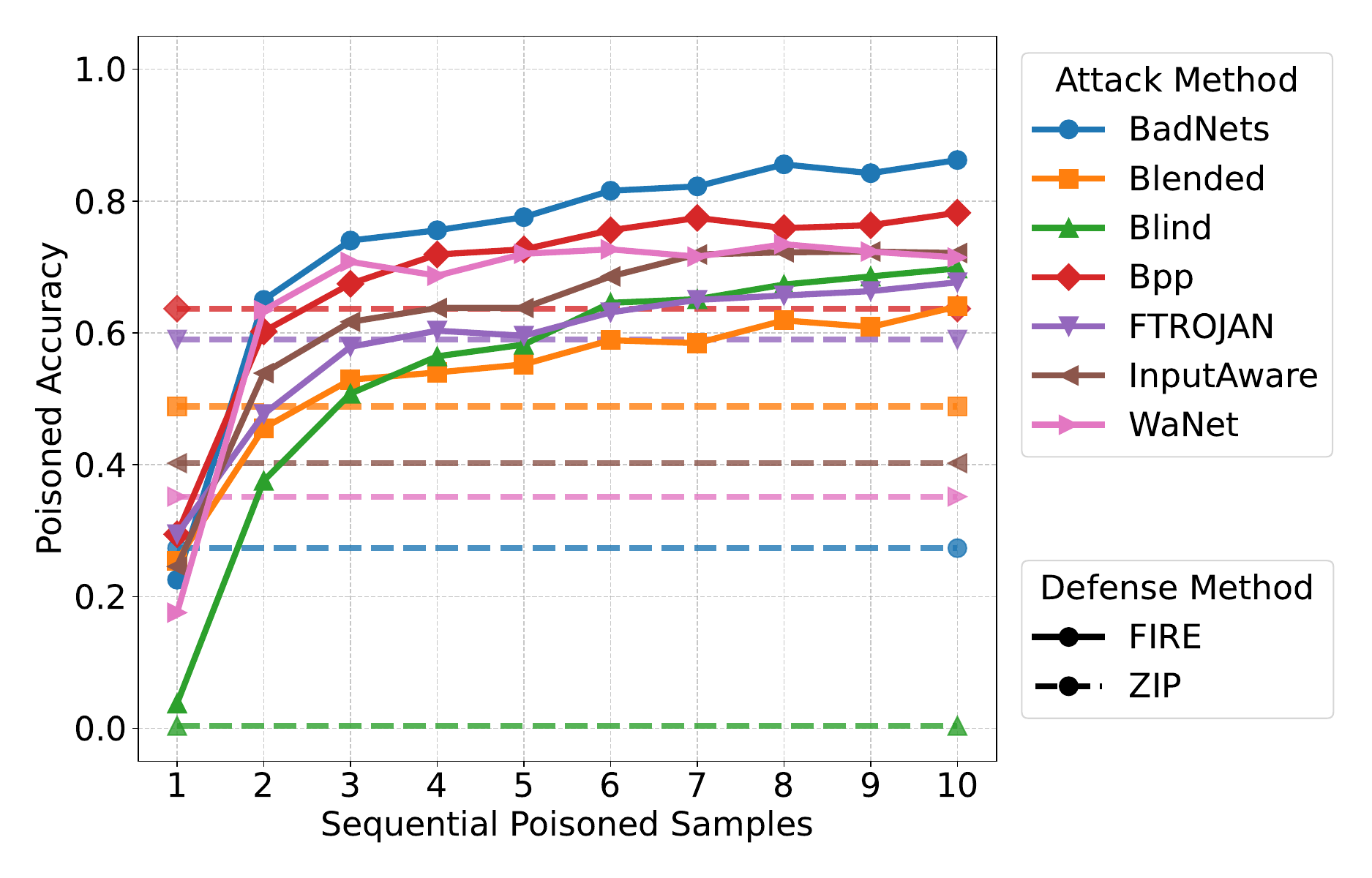}
  \caption{Runtime mitigation performance on CIFAR-10 with a PreActResNet18 using strategy 2 under a stream of poisoned samples. The x-axis denotes the poisoned sample index (arrival order), and the y-axis shows Poisoned Accuracy (PA, \%). \textit{FIRE} improves as additional poisoned samples arrive, whereas ZIP (dashed) remains constant because it does not adapt using new samples.}
  \label{fig:pa_cifar10_res18}
\end{figure}

Fig.~\ref{fig:pa_cifar10_res18} shows that \textit{FIRE} is particularly effective in the streaming setting: it updates its mitigation decision as additional poisoned samples arrive, leading to consistently improving PA over the stream. Across the full benchmark suite in Tab.~\ref{tab:runtime_results}, \textit{FIRE} already outperforms ZIP in \textbf{9/21} settings at the first poisoned sample (Pos~1), and it outperforms ZIP in \textbf{all 21/21} settings by the 10th poisoned sample (Pos~10). This trend is also reflected in Fig.~\ref{fig:pa_cifar10_res18}, where ZIP remains flat because it does not incorporate new poisoned samples, while \textit{FIRE} steadily improves and converges to high PA.

\subsection{Execution Time Analysis}
\label{subsec:execution_time_analysis}

In addition to mitigation effectiveness, practical test-time defenses must meet strict latency constraints. We therefore evaluate the execution time of \textit{FIRE} and ZIP in a \emph{streaming} setting, where samples arrive one after another and the defense is required to mitigate the input and output the corrected label as quickly as possible. All timing measurements were obtained on a single NVIDIA A100 GPU.

The right side of Tab.~\ref{tab:runtime_results} summarizes the runtime results. After a brief initialization of around $\sim$200--250\,ms, \textit{FIRE} enables consistently fast online processing across architectures, datasets, and attacks: it processes a sample in roughly \textbf{$\sim$11--27\,ms}, while ZIP requires about $\sim$1.4--1.6\,s per sample. This $\approx$two-order-of-magnitude speedup makes \textit{FIRE} well-suited for low-latency streaming inference, where real-time mitigation is essential.

\section{Conclusion and Limitations}
We analyzed backdoor behavior through intermediate representations and found that inspecting \emph{multiple} latent spaces of a poisoned model can reveal where the trigger effect becomes structured, and thus exploitable for mitigation. Across a diverse set of attacks, datasets, and architectures, there is typically at least one latent space in which the trigger manifests as a \emph{single, label-agnostic direction}, enabling reliable removal of the trigger component and recovery of the sample’s original label.
Building on this observation, we introduced \textit{FIRE} (\emph{Feature-space Inference-time REpair}), a lightweight runtime defense that repairs latent representations during a forward pass by estimating a backdoor direction and subtracting its component at an appropriate layer, without retraining or changing model parameters. Experiments show strong label recovery that improves as more poisoned samples are processed in a stream, while incurring substantially lower overhead than prior runtime correction baselines. At the same time, our evaluation is currently limited to image models; extending \textit{FIRE} to other domains (e.g., text or graphs) and to multi-objective backdoors that may require multiple shifts (possibly in multiple latent spaces) remains an important direction for future work.

\section{Impact Statement}
This paper presents work whose goal is to advance the field of Machine
Learning. There are many potential societal consequences of our work. However, it aims to improve the security of machine learning model usage, which benefits the community.

\bibliography{bibliography}
\bibliographystyle{icml2026}

\newpage
\appendix
\onecolumn

\section{Additional Strategies for Estimating the Trigger Direction}
\label{app:strategies}
\paragraph{\textit{FIRE} using augmentations only} Rather than relying on representations extracted from clean samples, the method can be formulated entirely in terms of image augmentations. Under this formulation, the initialization step, namely the computation of the clean centroid, is no longer required. Alg.~\ref{alg:runtime_mitigation_augment} presents this augmentation based variant.

\begin{figure*}[!b]
  \centering

  \begin{subfigure}[t]{0.305\textwidth}
    \centering
    \includegraphics[width=\linewidth]{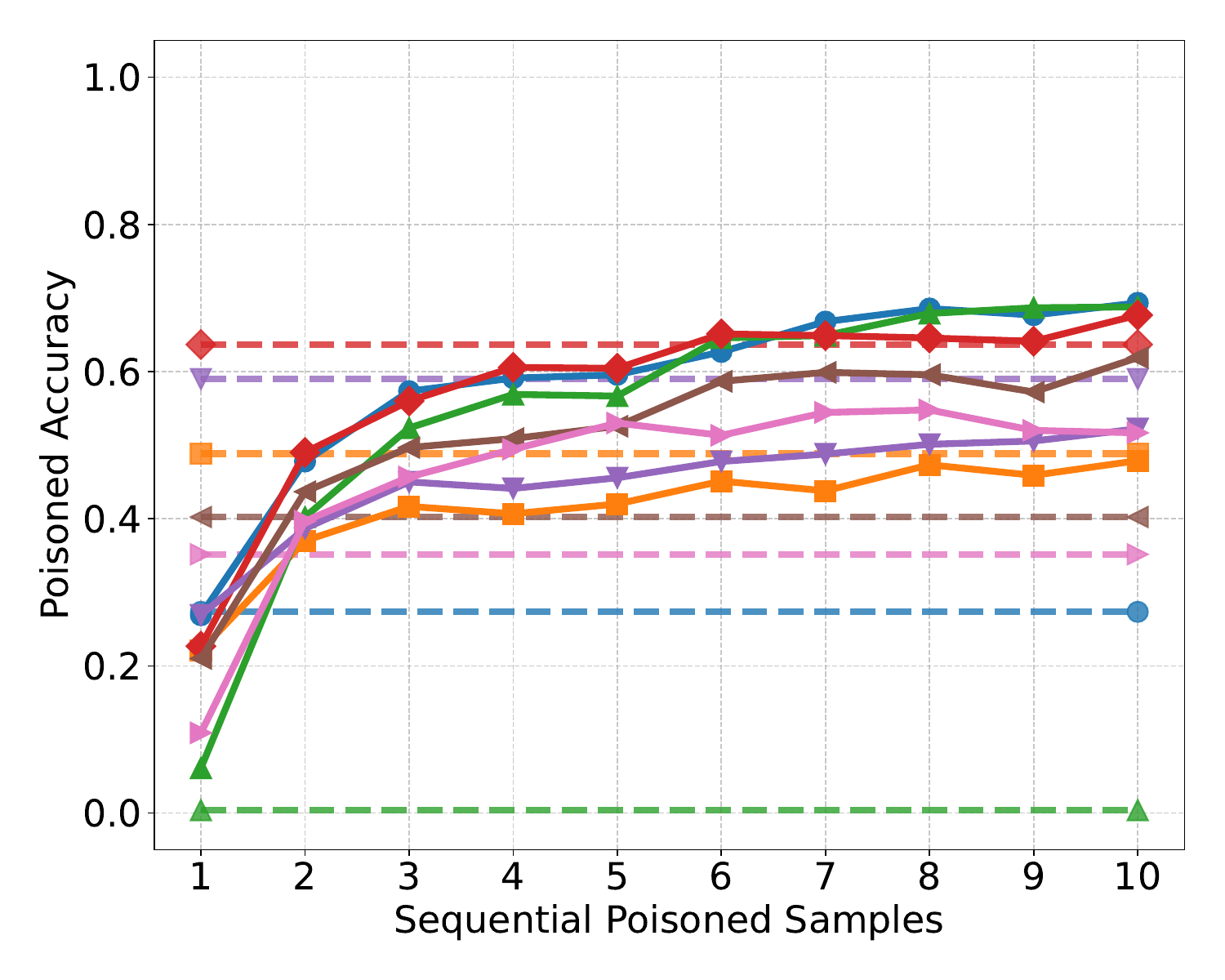}
    \caption{\textit{FIRE} with augmentations only}
    \label{fig:augment_only}
  \end{subfigure}\hfill
  \begin{subfigure}[t]{0.305\textwidth}
    \centering
    \includegraphics[width=\linewidth]{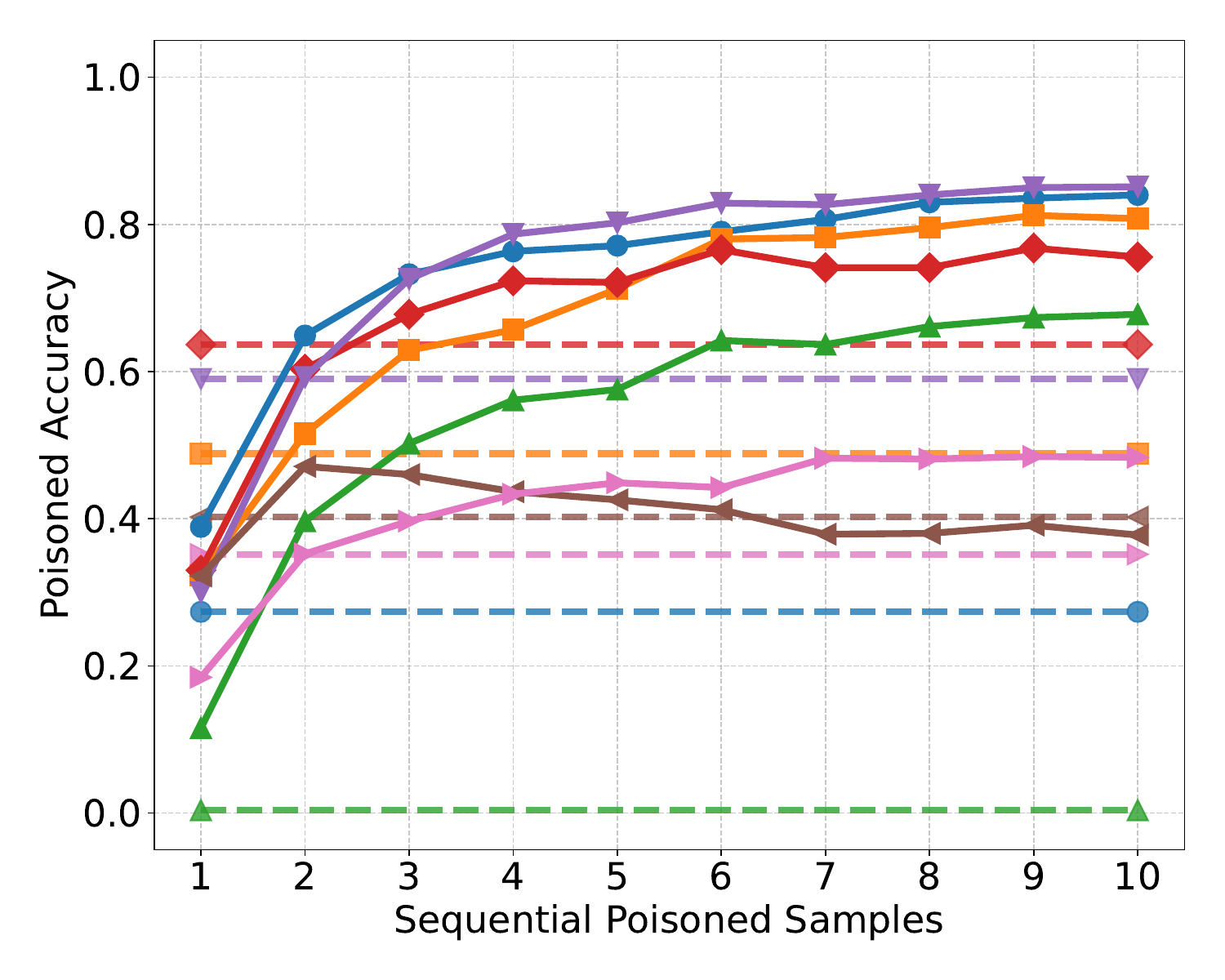}
    \caption{\textit{FIRE} with augmentations and projection}
    \label{fig:augment_only_proj}
  \end{subfigure}
  \begin{subfigure}[t]{0.38\textwidth}
    \centering
    \includegraphics[width=\linewidth]{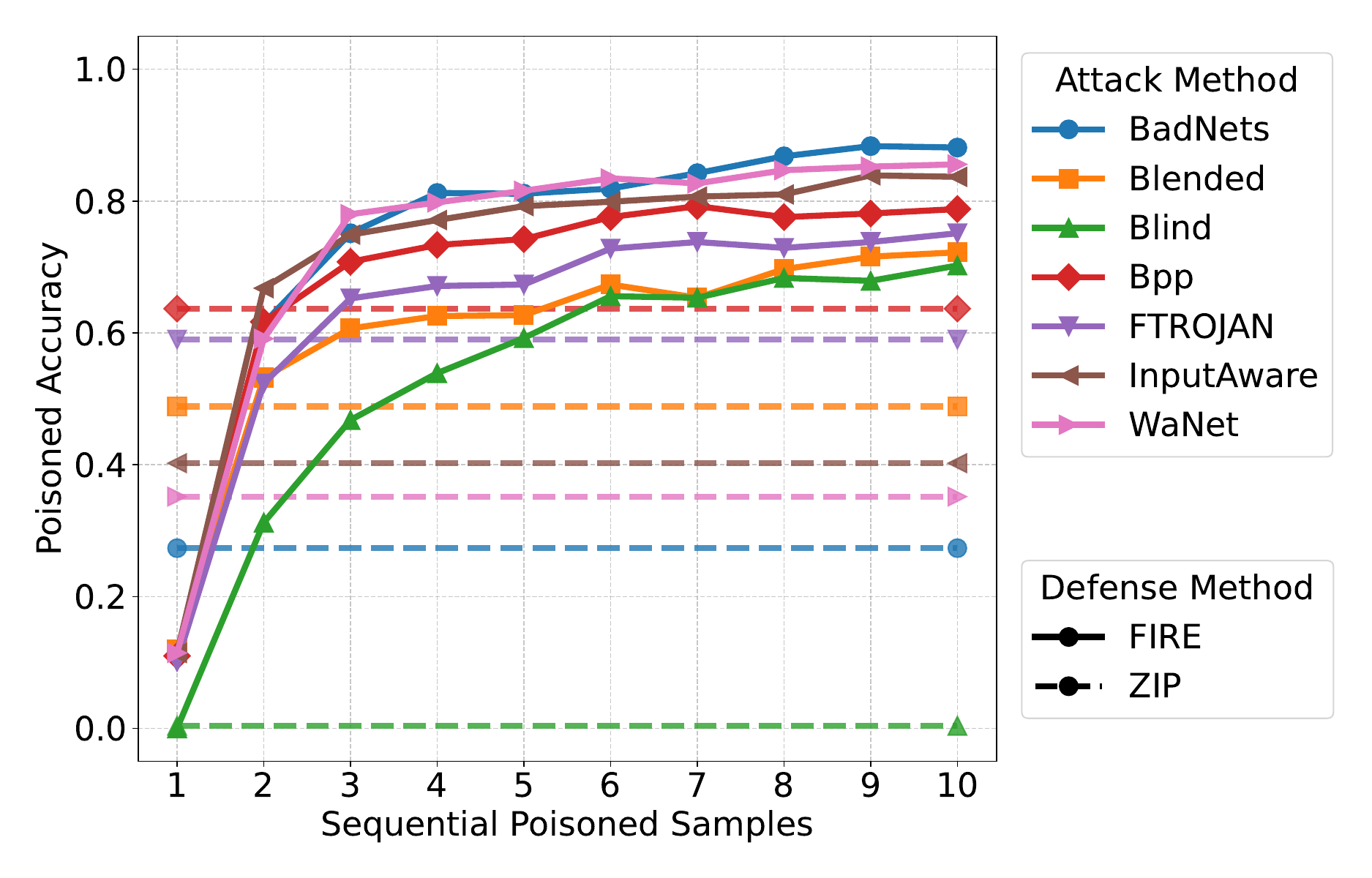}
    \caption{\textit{FIRE} without image augmentations}
    \label{fig:fire_basic}
  \end{subfigure}\hfill

  \caption{Mitigation performance on CIFAR-10 when using a PreActResNet18}
  \label{fig:fire_variants}
\end{figure*}

\begin{algorithm}[!b]
\caption{Runtime latent-space backdoor mitigation using augmentations only}
\label{alg:runtime_mitigation_augment}
\begin{algorithmic}
  \STATE \textbf{Input:} model $f$, candidate layers $\mathcal{L}$, augmentation $\mathcal{A}$
  \STATE $\widehat{\mu}^{\mathrm{pois,aug}}_{\ell}\leftarrow 0$ for all $\ell\in\mathcal{L}$
  \STATE Initialize counters $n_\ell \leftarrow 0$ and shifts $\widehat{b}_{\ell}\leftarrow 0$ for all $\ell\in\mathcal{L}$
  \FOR{each incoming poisoned sample $x_t^{\mathrm{pois}}$}
    \STATE $y\leftarrow \arg\max f(x_t^{\mathrm{pois}})$
    \STATE $x_t^{\mathrm{pois,aug}} \leftarrow \mathcal{A}\big(x_t^{\mathrm{pois}}\big)$
    \FOR{each $\ell\in\mathcal{L}$ in ascending order}
      \STATE $x_{t,\ell}^{\mathrm{pois}} \leftarrow h_\ell(x_t^{\mathrm{pois}})$,\quad $x_{t,\ell}^{\mathrm{pois,aug}} \leftarrow h_\ell(x_t^{\mathrm{pois,aug}})$,\quad $n_\ell \leftarrow n_\ell + 1$
      \STATE $\widehat{\mu}^{\mathrm{pois}}_\ell \leftarrow \widehat{\mu}^{\mathrm{pois}}_\ell + \frac{1}{n_\ell}\!\left(x_{t,\ell}^{\mathrm{pois}} - \widehat{\mu}^{\mathrm{pois}}_\ell\right)$
      \STATE $\widehat{\mu}^{\mathrm{pois,aug}}_\ell \leftarrow \widehat{\mu}^{\mathrm{pois,aug}}_\ell + \frac{1}{n_\ell}\!\left(x_{t,\ell}^{\mathrm{pois,aug}} - \widehat{\mu}^{\mathrm{pois,aug}}_\ell\right)$ 
      \STATE $\widehat{b}_\ell \leftarrow \widehat{\mu}^{\mathrm{pois}}-\widehat{\mu}^{\mathrm{pois,aug}}_\ell$
      \STATE $\tilde{x}_{t,\ell} \leftarrow \mathrm{Rep}_\ell\!\left(x_{t,\ell}^{\mathrm{pois}},\widehat{b}_\ell,1\right)$
      \STATE $y_{\mathrm{rep}} \leftarrow \arg\max g_\ell(\tilde{x}_{t,\ell})$
      \IF{$y_{\mathrm{rep}} \neq y$}
        \STATE \textbf{Output} $y_{\mathrm{rep}}$ and \textbf{break}
      \ENDIF
    \ENDFOR
    \STATE If no layer changed the prediction, \textbf{output} $y$
  \ENDFOR
\end{algorithmic}
\end{algorithm}
Fig.~\ref{fig:augment_only} reports the performance when using \textit{FIRE} using only image augmentations. As the number of samples increases, accuracy consistently improves, and even without additional techniques the model attains strong performance. While combining methods in Fig.~\ref{fig:pa_cifar10_res18} yields the better results, \textit{FIRE} with augmentations alone already provides a solid and effective baseline.

\paragraph{Projection-based Repair} If clean data is available, there is also the option of replacing the pure shift with a projection.
For a given sample, it removes only the component of $x_\ell$ that lies along the augmentation-based direction $\widehat{b}_\ell$ by projecting onto the affine subspace orthogonal to $\widehat{b}_\ell$ around the clean centroid $\widehat{\mu}^{\mathrm{clean}}_\ell$:
\begin{equation}
\tilde{x}_\ell
\;=\;
\mathrm{Rep}_\ell^{\mathrm{proj}}(x_\ell)
\;\coloneqq\;
x_\ell
-
\frac{\langle x_\ell-\widehat{\mu}^{\mathrm{clean}}_\ell,\, \widehat{b}_\ell\rangle}{\|\widehat{b}_\ell\|_2^2}\,\widehat{b}_\ell.
\label{eq:rep_project}
\end{equation}
This projection based repair adjusts $x_\ell$ only along the estimated trigger direction $\widehat{b}_\ell$, removing the $\widehat{b}_\ell$ aligned component of the displacement $x_\ell-\widehat{\mu}^{\mathrm{clean}}_\ell$. Equivalently, it maps $x_\ell$ to the unique point obtained by moving along $\widehat{b}_\ell$ that is closest in Euclidean distance to the clean centroid $\widehat{\mu}^{\mathrm{clean}}_\ell$, while leaving all components orthogonal to $\widehat{b}_\ell$ unchanged.

As shown in Fig.~\ref{fig:augment_only_proj}, projection-based repair improves accuracy for several attacks (BadNets, FTROJAN, Blended), while its effectiveness is attack-dependent and can degrade performance for others (InputAware, WaNet).

\begin{algorithm}[!t]
\caption{Runtime latent-space backdoor mitigation without augmentations}
\label{alg:runtime_mitigation_classic}
\begin{algorithmic}
  \STATE \textbf{Input:} model $f$, candidate layers $\mathcal{L}$, clean set $\mathcal{D}_{\mathrm{clean}}$
  \STATE Compute $\widehat{\mu}^{\mathrm{clean}}_\ell$ from $\mathcal{D}_{\mathrm{clean}}$ for all $\ell\in\mathcal{L}$ (Eq.~\ref{eq:clean_centroid_offline})
  \STATE Initialize $\widehat{\mu}^{\mathrm{pois}}_{\ell}\leftarrow 0$ for all $\ell\in\mathcal{L}$
  \STATE Initialize counters $n_\ell \leftarrow 0$ and shifts $\widehat{b}_{\ell}\leftarrow 0$ for all $\ell\in\mathcal{L}$

  \FOR{each incoming poisoned sample $x_t^{\mathrm{pois}}$}
    \STATE $y\leftarrow \arg\max f(x_t^{\mathrm{pois}})$
    \FOR{each $\ell\in\mathcal{L}$ in ascending order}
      \STATE $x_{t,\ell}^{\mathrm{pois}} \leftarrow h_\ell(x_t^{\mathrm{pois}})$,\quad $n_\ell \leftarrow n_\ell + 1$
      \STATE $\widehat{\mu}^{\mathrm{pois}}_\ell \leftarrow \widehat{\mu}^{\mathrm{pois}}_\ell + \frac{1}{n_\ell}\!\left(x_{t,\ell}^{\mathrm{pois}} - \widehat{\mu}^{\mathrm{pois}}_\ell\right)$
      \STATE $\widehat{b}_\ell \leftarrow \widehat{\mu}^{\mathrm{pois}}_\ell - \widehat{\mu}^{\mathrm{clean}}_\ell$
      \STATE $\tilde{x}_{t,\ell} \leftarrow \mathrm{Rep}_\ell\!\left(x_{t,\ell}^{\mathrm{pois}},\widehat{b}_\ell,1\right)$
      \STATE $y_{\mathrm{rep}} \leftarrow \arg\max g_\ell(\tilde{x}_{t,\ell})$
      \IF{$y_{\mathrm{rep}} \neq y$}
        \STATE \textbf{Output} $y_{\mathrm{rep}}$ and \textbf{break}
      \ENDIF
    \ENDFOR
    \STATE If no layer changed the prediction, \textbf{output} $y$
  \ENDFOR
\end{algorithmic}
\end{algorithm}

\paragraph{\textit{FIRE} without image augmentations}
Li et al.\ show that, when a backdoored network is trained with image augmentations, it can become invariant to these transformations \yrcite{liBackdoorAttackPhysical2021}, which can reduce the effectiveness of augmentation based variants. Nevertheless, \textit{FIRE} can also be applied without augmentations, as summarized in Alg.~\ref{alg:runtime_mitigation_classic}.

Without augmentations, estimating the repair direction from a single incoming sample yields a trivial estimate: the resulting direction is simply the vector from the clean centroid to that sample in latent space. Consequently, repairing the first sample shifts it directly to the clean centroid, and the resulting prediction becomes largely independent of the sample's original position. This limitation is visible in Fig.~\ref{fig:fire_basic}. In the Blind attack setting, the centroid is (incidentally) classified as the target class, which leads to a PA of $0$ for the first repaired sample. As more samples are observed, however, the trigger direction can be approximated more reliably, and subsequent samples can be repaired meaningfully.

\section{Comparison with ShrinkPad}
\label{app:shrinkpad}
Besides ZIP, we also evaluated ShrinkPad~\cite{liBackdoorAttackPhysical2021}. ShrinkPad is an input-space purification baseline that shrinks and pads the input image to disrupt potential trigger patterns.

Tab.~\ref{tab:runtime_results_shrinkpad} summarizes results in the streaming setting. Overall, ShrinkPad is consistently fast (online runtime of roughly $3$--$5$\,ms per image), but its PA varies substantially across attacks and datasets. In contrast, \textit{FIRE} achieves markedly higher PA in most configurations, especially at \textit{Pos 10}, indicating that the latent-space correction benefits from a longer observation window. This robustness comes with a higher online cost (about $12$--$26$\,ms) and an additional one-time initialization (about $205$--$241$\,ms), but remains practical for streaming deployment.

Notably, there are a few cases where ShrinkPad matches or slightly exceeds \textit{FIRE} (e.g., FTROJAN in several settings), suggesting that simple input transformations can be effective when the trigger is highly sensitive to resizing. However, across the broader benchmark, \textit{FIRE} provides a more reliable PA improvement while maintaining competitive runtime.

\begin{table*}[t]
\caption{Benchmark results for the streaming scenario: Clean Accuracy (CA, higher is better), Poisoned Accuracy (PA, higher is better) for the ShrinkPad and \textit{FIRE} approaches, and the Time in milliseconds (ms, lower is better) are reported.}
\label{tab:runtime_results_shrinkpad}
\vskip 0.1in
\begin{center}
\begin{small}
\begin{sc}
\begin{tabular}{llcc|ccc|ccc}
\toprule
Configuration & Attack & \multicolumn{2}{c}{No defense} & ShrinkPad & \multicolumn{2}{c}{FIRE} & ShrinkPad & \multicolumn{2}{c}{FIRE} \\
\cmidrule(lr){3-4} \cmidrule(lr){5-5} \cmidrule(lr){6-7} \cmidrule(lr){8-8} \cmidrule(lr){9-10}
 & & CA & PA & PA $\uparrow$& \multicolumn{2}{c}{PA $\uparrow$} & Time  $\downarrow$& \multicolumn{2}{c}{Time $\downarrow$} \\
\cmidrule(lr){6-7} \cmidrule(lr){9-10}
 & & & & & Pos 1 & Pos 10 & Online & Init. & Online \\
\midrule
 & BadNets & 91.86 & 5.27 & 67.43 & 22.56 & \textbf{86.22} & \textbf{3.0} & 223.7 & 14.0 \\
 & Blended & 93.57 & 0.18 & 33.82 & 25.44 & \textbf{64.00} & \textbf{2.8} & 222.0 & 12.1 \\
 & Blind & 80.20 & 0.40 & 26.94 & 3.78 & \textbf{69.78} & \textbf{2.9} & 228.5 & 15.2 \\
\smash{\raisebox{-0.5\normalbaselineskip}{\shortstack{PreActResNet18 \\ CIFAR-10}}} & Bpp & 91.07 & 0.39 & 75.86 & 29.44 & \textbf{78.22} & \textbf{3.1} & 213.6 & 11.8 \\
 & FTROJAN & 93.56 & 0.00 & \textbf{78.89} & 29.44 & 67.67 & \textbf{2.7} & 220.4 & 12.0 \\
 & InputAware & 91.29 & 2.54 & 62.41 & 24.56 & \textbf{72.11} & \textbf{3.0} & 214.4 & 12.1 \\
 & WaNet & 91.24 & 3.90 & 65.63 & 17.56 & \textbf{71.44} & \textbf{2.9} & 219.2 & 16.8 \\
\midrule
 & BadNets & 96.05 & 3.73 & 27.06 & 76.53 & \textbf{86.95} & \textbf{2.8} & 225.8 & 11.7 \\
 & Blended & 97.06 & 0.27 & 4.14 & 34.61 & \textbf{61.89} & \textbf{2.8} & 231.8 & 12.8 \\
 & Blind & 57.33 & 41.23 & 33.69 & 21.88 & \textbf{49.01} & \textbf{2.9} & 225.5 & 11.5 \\
\smash{\raisebox{-0.5\normalbaselineskip}{\shortstack{PreActResNet18 \\ GTSRB}}} & Bpp & 96.80 & 0.24 & 28.20 & \textbf{85.28} & 82.66 & \textbf{3.0} & 228.2 & 11.7 \\
 & FTROJAN & 98.12 & 0.00 & \textbf{91.73} & 79.16 & 76.37 & \textbf{3.0} & 241.0 & 12.0 \\
 & InputAware & 97.53 & 6.17 & 58.50 & \textbf{71.20} & 61.89 & \textbf{3.0} & 232.6 & 12.2 \\
 & WaNet & 96.79 & 7.06 & 41.69 & 20.29 & \textbf{42.32} & \textbf{3.0} & 226.2 & 14.2 \\
\midrule
 & BadNets & 92.46 & 4.31 & 64.20 & 26.78 & \textbf{88.67} & \textbf{5.0} & 208.5 & 22.1 \\
 & Blended & 94.02 & 0.21 & 37.56 & 26.89 & \textbf{61.78} & \textbf{5.3} & 205.5 & 20.4 \\
 & Blind & 89.43 & 0.03 & 48.64 & 19.78 & \textbf{78.44} & \textbf{5.0} & 212.1 & 24.2 \\
\smash{\raisebox{-0.5\normalbaselineskip}{\shortstack{PreActResNet34 \\ CIFAR-10}}} & Bpp & 92.23 & 0.28 & 65.51 & 25.44 & \textbf{71.11} & \textbf{5.2} & 209.4 & 21.8 \\
 & FTROJAN & 93.63 & 0.06 & 78.13 & 29.11 & \textbf{82.89} & \textbf{5.0} & 210.8 & 21.1 \\
 & InputAware & 91.61 & 3.71 & 59.36 & 21.11 & \textbf{71.33} & \textbf{5.0} & 221.1 & 23.2 \\
 & WaNet & 90.17 & 9.74 & 49.66 & 13.33 & \textbf{50.78} & \textbf{4.9} & 209.5 & 26.2 \\
\bottomrule
\end{tabular}
\end{sc}
\end{small}
\end{center}
\vskip -0.1in
\end{table*}

\begin{figure}[!t]
  \centering
  \includegraphics[width=0.5\columnwidth]{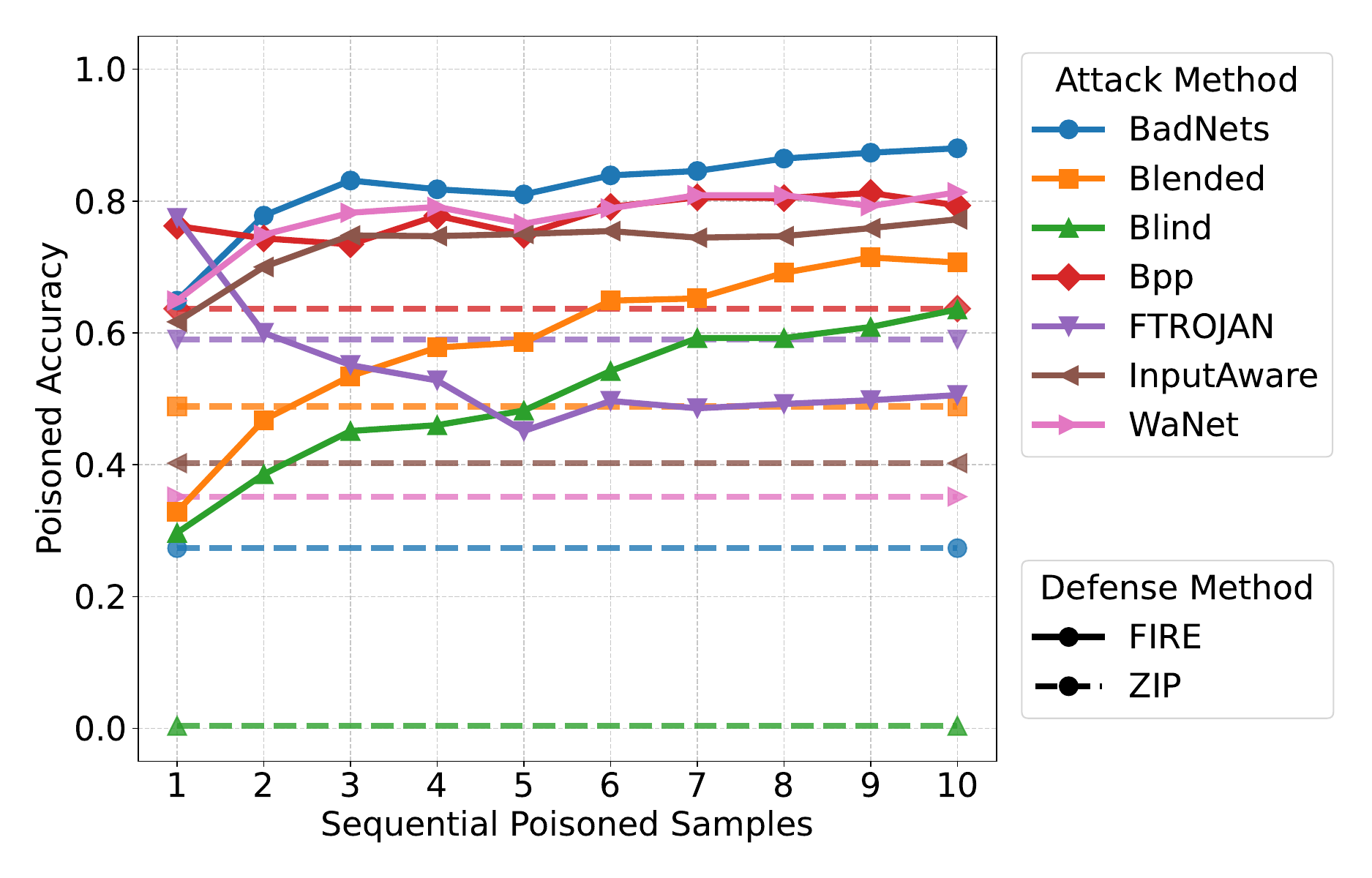}
  \caption{Mitigation performance on CIFAR-10 when using a PreActResNet18 when \textit{FIRE} uses modified images generated by ShrinkPad as guidance.}
  \label{fig:shrinkpad+fire}
\end{figure}

Additionally, we evaluated ShrinkPad as an augmentation technique in combination with \textit{FIRE} using Alg.~\ref{alg:runtime_mitigation_augment}. As shown in Fig.~\ref{fig:shrinkpad+fire}, for the first incoming poisoned sample the PA matches that of ShrinkPad alone, and for $6$ out of $7$ attacks performance improves as more examples are observed.

\section{Training Details for the Backdoored Models}
\label{app:training_details}
Table~\ref{tab:attack_details} summarizes the training configurations and resulting performance metrics for the backdoored models used in Section~\ref{sec:proof-of-concept} and \ref{sec:experiments}. All models were trained following the default protocol and hyperparameter settings provided by BackdoorBench~\cite{wuBackdoorBenchComprehensiveBenchmark2025}. For each attack, we report the number of training epochs (Epochs), CA, the attack success rate under the trigger (ASR), and PA.

Overall, most attack configurations achieve high CA while attaining strong ASR, indicating effective backdoor implantation with limited degradation of clean performance. One notable exception is the Blind attack on GTSRB: the resulting model exhibits unusually low CA (57.28\%) and comparatively low ASR (53.28\%), alongside a markedly elevated PA (41.23\%). This pattern suggests that the model did not converge to a well-performing solution under this configuration, i.e., Blind failed to train properly on GTSRB in our setting.

\begin{table*}[t]
\caption{Training details for each of the attacks: Number of training epochs (Epochs), Clean Accuracy (CA), Attack Success Rate (ASR), and Poisoned Accuracy (PA) are reported.}
\label{tab:attack_details}
\vskip 0.1in
\begin{center}
\begin{small}
\begin{sc}
\begin{tabular}{llcccc}
\toprule
Configuration & Attack & Epochs & CA & ASR & PA \\
\midrule
 & BadNets & 100 & 91.84 & 94.52 & 5.27 \\
 & Blended & 100 & 93.57 & 99.81 & 0.18 \\
 & Blind & 100 & 80.17 & 99.59 & 0.40 \\
\smash{\raisebox{-0.5\normalbaselineskip}{\shortstack{PreActResNet18 \\ CIFAR-10}}} & Bpp & 100 & 90.99 & 99.61 & 0.39 \\
 & FTROJAN & 100 & 93.56 & 100.00 & 0.00 \\
 & InputAware & 100 & 91.28 & 97.36 & 2.54 \\
 & WaNet & 100 & 91.19 & 95.83 & 3.90 \\
\midrule
 & BadNets & 50 & 96.05 & 96.13 & 3.73 \\
 & Blended & 50 & 97.05 & 99.71 & 0.27 \\
 & Blind & 50 & 57.28 & 53.28 & 41.23 \\
\smash{\raisebox{-0.5\normalbaselineskip}{\shortstack{PreActResNet18 \\ GTSRB}}} & Bpp & 50 & 96.79 & 99.76 & 0.24 \\
 & FTROJAN & 50 & 98.11 & 100.00 & 0.00 \\
 & InputAware & 50 & 97.51 & 93.38 & 6.17 \\
 & WaNet & 50 & 96.80 & 92.46 & 7.06 \\
\midrule
 & BadNets & 100 & 92.47 & 95.43 & 4.31 \\
 & Blended & 100 & 94.04 & 99.79 & 0.21 \\
 & Blind & 100 & 89.36 & 99.97 & 0.03 \\
\smash{\raisebox{-0.5\normalbaselineskip}{\shortstack{PreActResNet34 \\ CIFAR-10}}} & Bpp & 100 & 92.21 & 99.72 & 0.28 \\
 & FTROJAN & 100 & 93.60 & 99.94 & 0.06 \\
 & InputAware & 100 & 91.57 & 96.13 & 3.71 \\
 & WaNet & 100 & 90.15 & 89.67 & 9.74 \\
\bottomrule
\end{tabular}
\end{sc}
\end{small}
\end{center}
\vskip -0.1in
\end{table*}

\section{Impact of Imperfect Poisoned Sample Detection}
\label{app:imperfect_detection}
In Section~\ref{sec:input_stream}, we assume access to a stream of detected poisoned inputs that is provided to \textit{FIRE}. In practice, however, any detector is likely to be imperfect and may occasionally misclassify clean samples as poisoned. We therefore evaluate how such false positives affect the mitigation performance of \textit{FIRE}.

\begin{figure*}[t]
  \centering
  \begin{subfigure}[t]{0.4\textwidth}
    \centering
    \includegraphics[width=\linewidth]{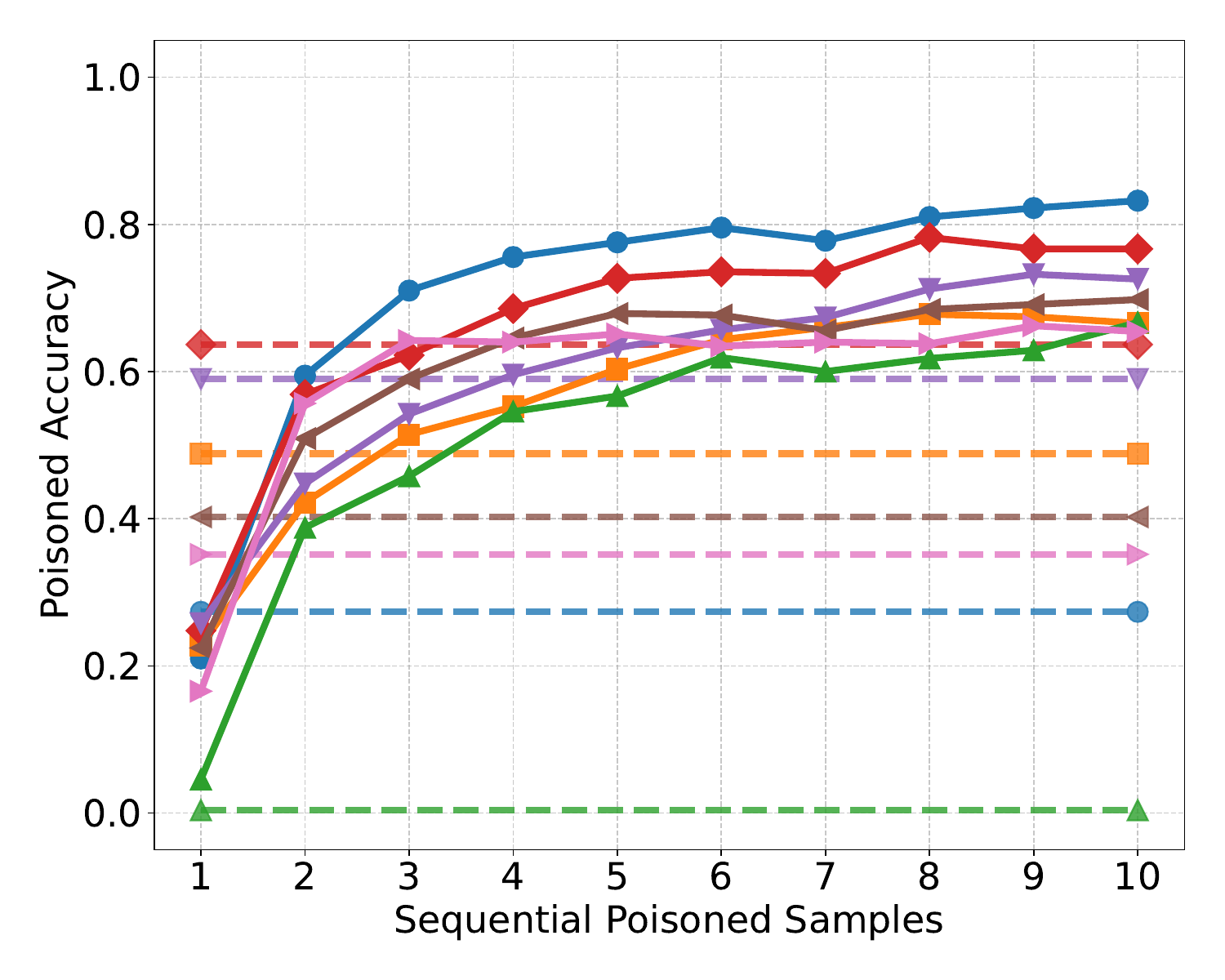}
    \caption{90\% poisoned samples, 10\% clean samples}
    \label{fig:fire_imperfect_9}
  \end{subfigure}\hfill
  \begin{subfigure}[t]{0.505\textwidth}
    \centering
    \includegraphics[width=\linewidth]{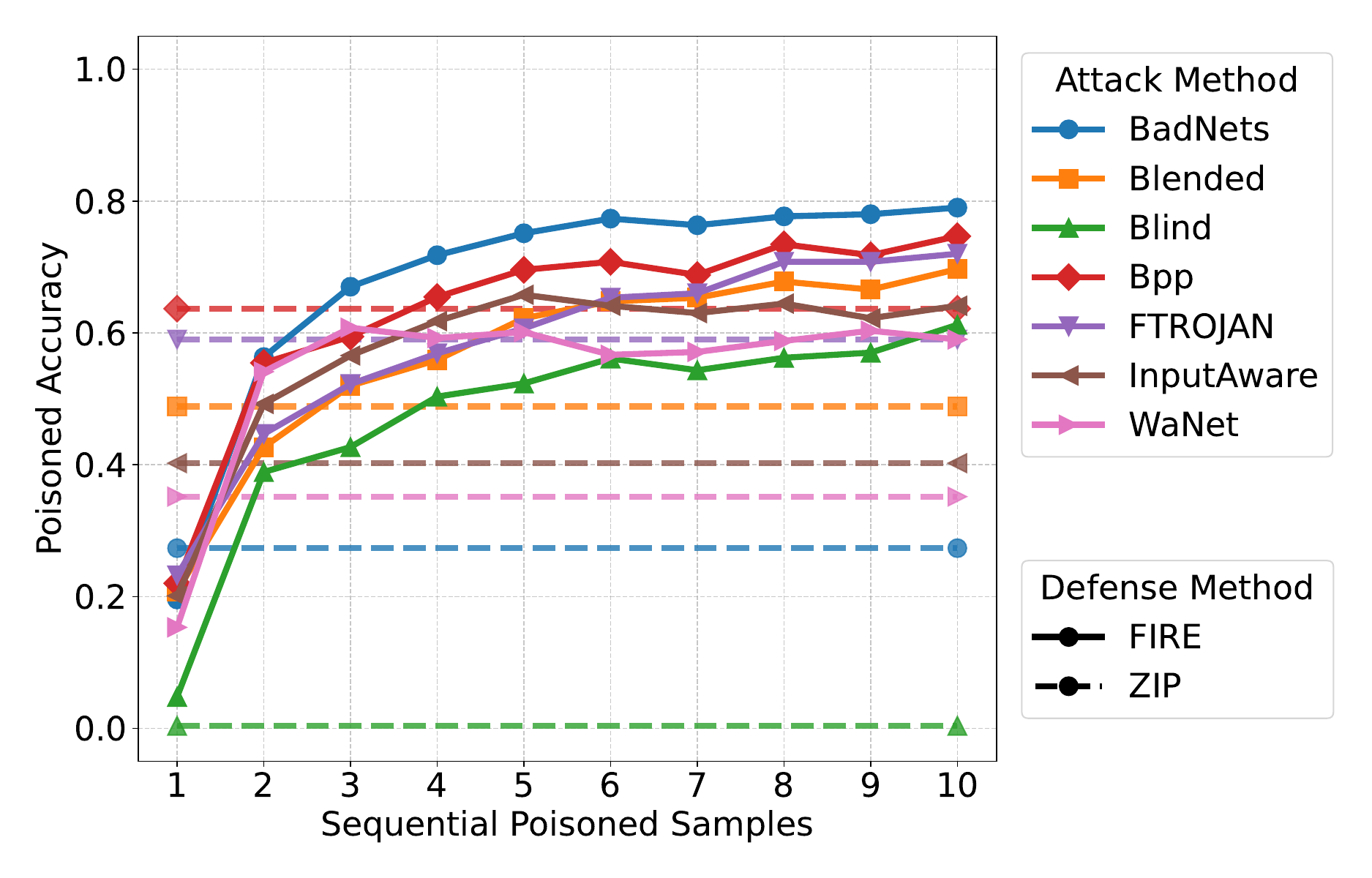}
    \caption{80\% poisoned samples, 20\% clean samples}
    \label{fig:fire_imperfect_8}
  \end{subfigure}
  \caption{Mitigation performance on CIFAR-10 when using a PreActResNet18 when the detection is imperfect.}
  \label{fig:fire_imperfect_all}
\end{figure*}

As illustrated in Fig.~\ref{fig:fire_imperfect_all}, increasing the fraction of incorrectly flagged clean samples leads to a gradual reduction in performance, as expected. Crucially, \textit{FIRE} remains effective even under these imperfect conditions: across both settings, the method continues to mitigate the attack and maintains strong PA throughout the stream. Overall, these results indicate that \textit{FIRE} is resilient to realistic detector noise and can still provide reliable protection when poisoned-sample detection is not fully accurate.

\begin{table*}[t]
\caption{Benchmark results comparing \textit{FIRE} performance with imperfect detectors. Poisoned Accuracy (PA, \%) is reported at Position 1 and Position 10. Differences in PA at Position 10 are also shown.}
\label{tab:fire_imperfect_comparison}
\vskip 0.1in
\begin{center}
\begin{small}
\begin{sc}
\begin{tabular}{llccccccccc}
\toprule
 & & \multicolumn{2}{c}{FIRE PA $\uparrow$} & \multicolumn{2}{c}{FIRE PA $\uparrow$ (90\%)} & \multicolumn{2}{c}{FIRE PA $\uparrow$ (80\%)} & \multicolumn{2}{c}{Difference (Pos 10)} \\
\cmidrule(lr){3-4} \cmidrule(lr){5-6} \cmidrule(lr){7-8} \cmidrule(lr){9-10}
Configuration & Attack & Pos 1 & Pos 10 & Pos 1 & Pos 10 & Pos 1 & Pos 10 & 0.9 - Std & 0.8 - Std \\
\midrule
 & BadNets & \textbf{22.56} & \textbf{86.22} & 21.00 & 83.22 & 19.56 & 79.00 & -3.00 & -7.22 \\
 & Blended & \textbf{25.44} & 64.00 & 22.78 & 66.56 & 20.78 & \textbf{69.67} & 2.56 & 5.67 \\
 & Blind & 3.78 & \textbf{69.78} & 4.56 & 66.67 & \textbf{4.78} & 61.22 & -3.11 & -8.56 \\
\smash{\raisebox{-0.5\normalbaselineskip}{\shortstack{PreActResNet18 \\ CIFAR-10}}} & Bpp & \textbf{29.44} & \textbf{78.22} & 24.78 & 76.67 & 22.00 & 74.67 & -1.55 & -3.55 \\
 & FTROJAN & \textbf{29.44} & 67.67 & 25.78 & \textbf{72.56} & 23.22 & 72.00 & 4.89 & 4.33 \\
 & InputAware & \textbf{24.56} & \textbf{72.11} & 22.44 & 69.78 & 20.11 & 64.11 & -2.33 & -8.00 \\
 & WaNet & \textbf{17.56} & \textbf{71.44} & 16.56 & 65.44 & 15.33 & 59.00 & -6.00 & -12.44 \\
\midrule
 & Average & \textbf{21.83} & \textbf{72.78} & 19.70 & 71.56 & 17.97 & 68.52 & -1.22 & -4.25 \\
\bottomrule
\end{tabular}
\end{sc}
\end{small}
\end{center}
\vskip -0.1in
\end{table*}

Tab.~\ref{tab:fire_imperfect_comparison} complements this observation with a quantitative comparison. Even when the detector stream contains a non-trivial fraction of clean samples (90\% / 80\% poisoned), \textit{FIRE} achieves PA that remains close to the standard setting in most cases. While some attacks incur moderate drops at Position~10 (e.g., BadNets and WaNet), others are largely unaffected, suggesting that \textit{FIRE} does not critically depend on perfectly filtered inputs. On average, the PA at Position~10 decreases only by $1.22$ points for the 90\% setting and $4.25$ points for the 80\% setting, underscoring that \textit{FIRE} continues to perform robustly under imperfect detection.

\section{Impact of the Number of Clean Samples on Direction Estimation}
\label{app:clean_samples}
\begin{figure}[!t]
  \centering
  \includegraphics[width=0.5\columnwidth]{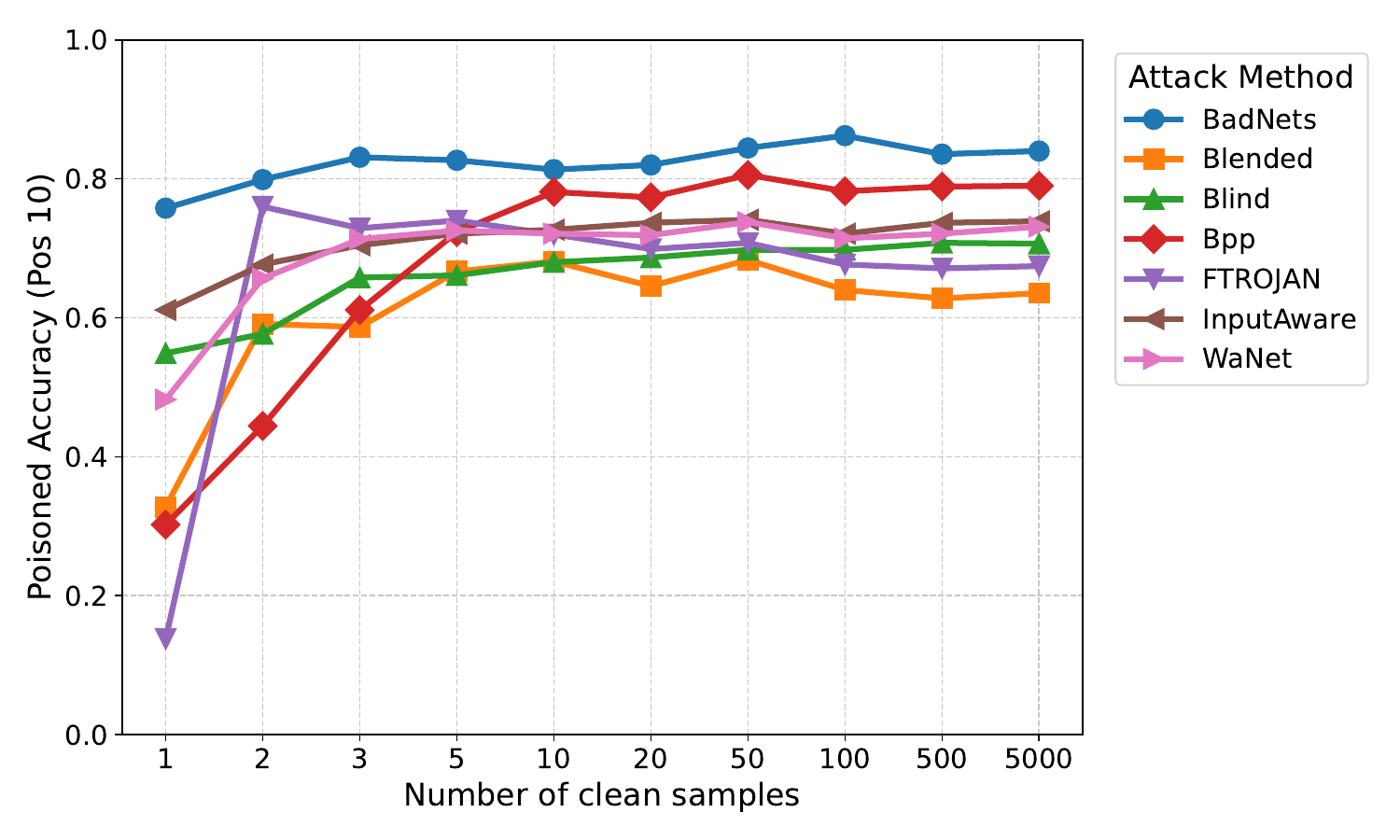}
  \caption{Poisoned accuracy for the $10$th processed sample on CIFAR-10 using a PreActResNet18, as a function of the number of clean samples used to estimate the clean centroid.}
  \label{fig:clean_sample_impact}
\end{figure}

Fig.~\ref{fig:clean_sample_impact} reports the PA for the $10$th processed sample as a function of the number of clean samples used to estimate the clean centroid. Even with very limited clean data, \textit{FIRE} achieves competitive performance, and increasing the number of clean samples rapidly strengthens the centroid estimate. For most attacks, the PA stabilizes once roughly $10$ clean samples are available, indicating that only a small clean set is sufficient in practice. Overall, these results highlight that \textit{FIRE} is sample-efficient and remains effective when clean data access is constrained.

\end{document}